\pdfoutput=1
\documentclass[11pt]{article}

\usepackage[]{acl}

\usepackage{times}
\usepackage{latexsym}

\usepackage[T1]{fontenc}
\usepackage[utf8]{inputenc}

\usepackage{microtype}

\usepackage{inconsolata}

\usepackage{microtype}
\usepackage{booktabs} % for professional tables
\usepackage{hyperref}
\usepackage{amsmath}
\usepackage{amssymb}
\usepackage{mathtools}
\usepackage{amsthm}
\usepackage{amsmath}
\usepackage{amssymb}
\usepackage{mathtools}
\usepackage{amsthm}
\usepackage{makecell}
\usepackage{bbm}
\usepackage[textsize=tiny]{todonotes}
\usepackage[capitalize,noabbrev]{cleveref}
\usepackage{inconsolata}
\usepackage{pythonhighlight}
\usepackage{algorithm2e}
\usepackage{utfsym}
\usepackage{fontawesome}
\usepackage{multirow}
\usepackage{booktabs}
\usepackage{adjustbox}
\usepackage{enumitem}
\usepackage{graphicx}
\usepackage{subfigure}
\usepackage{xcolor}
\usepackage{colortbl}
\usepackage{pgf}
\usepackage{pgfplotstable} % For handling and displaying tables
\usepackage{pgfplots} % For calculations and color mappings
\usepackage{colortbl} % For coloring table cells
\usepackage{booktabs} % For better-looking tables
\usepackage{etoolbox}
\usepackage{twemojis}

\newcommand*\opensource{
    {\centering
\includegraphics[width=0.3cm]{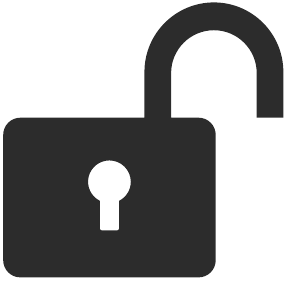}}}
\newcommand*\closedsource{
{\centering\includegraphics[width=0.25cm]{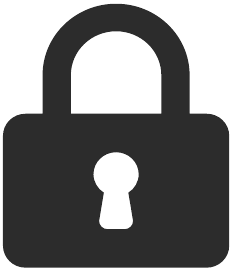}}}

\definecolor{ColorLow}{HTML}{BEE7A5} % Light green
\definecolor{ColorMid}{HTML}{FFEDA0} % Light yellow
\definecolor{ColorHigh}{HTML}{FFC0CB} % Light pink

\newcommand{\applyGradient}[1]{
    \ifdim #1 pt > 50 pt
    \pgfmathparse{((#1 - 46.5) / 11) * 100}
    \let\Lightness\pgfmathresult
    \edef\temp{\noexpand\cellcolor{ColorHigh!\Lightness!white}}\temp
    \else
        \ifdim #1 pt > 40 pt
            \pgfmathparse{((#1 - 38.2) / 11) * 100}
                \let\Lightness\pgfmathresult
                \edef\temp{\noexpand\cellcolor{ColorMid!\Lightness!white}}\temp        \else
                \pgfmathparse{((#1 - 25) / 14) * 100}
    \let\Lightness\pgfmathresult
            \edef\temp{\noexpand\cellcolor{ColorLow!\Lightness!white}}\temp % Lower values get lighter color
        \fi
    \fi
    #1
}

\newcommand{\applyGradienta}[1]{
    \ifdim #1 pt > 50 pt
    \pgfmathparse{((#1 - 46.5) / 10) * 100}
    \let\Lightness\pgfmathresult
    \edef\temp{\noexpand\cellcolor{ColorHigh!\Lightness!white}}\temp
    \else
        \ifdim #1 pt > 40 pt
            \pgfmathparse{((#1 - 38.2) / 11) * 100}
                \let\Lightness\pgfmathresult
                \edef\temp{\noexpand\cellcolor{ColorMid!\Lightness!white}}\temp        \else
                \pgfmathparse{((#1 - 30) / 8) * 100}
    \let\Lightness\pgfmathresult
            \edef\temp{\noexpand\cellcolor{ColorLow!\Lightness!white}}\temp % Lower values get lighter color
        \fi
    \fi
    #1
}

\newcommand{\applyGradientb}[1]{
 \ifdefstring{#1}{-}{%
        \textit{#1} % Example: italicize the dash
    }{%
    \ifdim #1 pt > 50 pt
    \pgfmathparse{((#1 - 46.5) / 10) * 100}
    \let\Lightness\pgfmathresult
    \edef\temp{\noexpand\cellcolor{ColorHigh!\Lightness!white}}\temp
    \else
        \ifdim #1 pt > 41 pt
            \pgfmathparse{((#1 - 38.2) / 11) * 100}
                \let\Lightness\pgfmathresult
                \edef\temp{\noexpand\cellcolor{ColorMid!\Lightness!white}}\temp        \else
                \pgfmathparse{((#1 - 10) / 16) * 100}
    \let\Lightness\pgfmathresult
            \edef\temp{\noexpand\cellcolor{ColorLow!\Lightness!white}}\temp % Lower values get lighter color
        \fi
    \fi
    #1
    }
}

\newcommand{\applyGradientc}[1]{
    \ifdim #1 pt > 50 pt
    \pgfmathparse{((#1 - 46.5) / 18) * 100}
    \let\Lightness\pgfmathresult
    \edef\temp{\noexpand\cellcolor{ColorHigh!\Lightness!white}}\temp
    \else
        \ifdim #1 pt > 40 pt
            \pgfmathparse{((#1 - 38.2) / 13) * 100}
                \let\Lightness\pgfmathresult
                \edef\temp{\noexpand\cellcolor{ColorMid!\Lightness!white}}\temp        \else
                \pgfmathparse{((#1 - 24) / 17) * 100}
    \let\Lightness\pgfmathresult
            \edef\temp{\noexpand\cellcolor{ColorLow!\Lightness!white}}\temp % Lower values get lighter color
        \fi
    \fi
    #1
}

\newcommand{\applyGradientd}[1]{
    \ifdim #1 pt > 40 pt
    \pgfmathparse{((#1 - 40) / 10) * 100}
    \let\Lightness\pgfmathresult
    \edef\temp{\noexpand\cellcolor{ColorHigh!\Lightness!white}}\temp
    \else
        \ifdim #1 pt > 30 pt
            \pgfmathparse{((#1 - 30) / 10) * 100}
                \let\Lightness\pgfmathresult
                \edef\temp{\noexpand\cellcolor{ColorMid!\Lightness!white}}\temp        \else
                \pgfmathparse{((#1 - 10) / 20) * 100}
    \let\Lightness\pgfmathresult
            \edef\temp{\noexpand\cellcolor{ColorLow!\Lightness!white}}\temp % Lower values get lighter color
        \fi
    \fi
    #1
}

\newcommand{\applyGradiente}[1]{
    \ifdim #1 pt > 80 pt
    \pgfmathparse{((#1 - 80) / 20) * 100}
    \let\Lightness\pgfmathresult
    \edef\temp{\noexpand\cellcolor{ColorHigh!\Lightness!white}}\temp
    \else
        \ifdim #1 pt > 60 pt
            \pgfmathparse{((#1 - 60) / 20) * 100}
                \let\Lightness\pgfmathresult
                \edef\temp{\noexpand\cellcolor{ColorMid!\Lightness!white}}\temp        \else
                \pgfmathparse{((#1 - 30) / 30) * 100}
    \let\Lightness\pgfmathresult
            \edef\temp{\noexpand\cellcolor{ColorLow!\Lightness!white}}\temp % Lower values get lighter color
        \fi
    \fi
    #1
}

\newcommand{\applyGradientf}[1]{
    \ifdim #1 pt > 70 pt
    \pgfmathparse{((#1 - 70) / 30) * 100}
    \let\Lightness\pgfmathresult
    \edef\temp{\noexpand\cellcolor{ColorHigh!\Lightness!white}}\temp
    \else
        \ifdim #1 pt > 40 pt
            \pgfmathparse{((#1 - 40) / 30) * 100}
                \let\Lightness\pgfmathresult
                \edef\temp{\noexpand\cellcolor{ColorMid!\Lightness!white}}\temp        \else
                \pgfmathparse{((#1 - 7) / 33) * 100}
    \let\Lightness\pgfmathresult
            \edef\temp{\noexpand\cellcolor{ColorLow!\Lightness!white}}\temp % Lower values get lighter color
        \fi
    \fi
    #1
}

\newcommand{\applyGradientg}[1]{
    \ifdim #1 pt > 40 pt
    \pgfmathparse{((#1 - 40) / 8) * 100}
    \let\Lightness\pgfmathresult
    \edef\temp{\noexpand\cellcolor{ColorHigh!\Lightness!white}}\temp
    \else
        \ifdim #1 pt > 20 pt
            \pgfmathparse{((#1 - 20) / 10) * 100}
                \let\Lightness\pgfmathresult
                \edef\temp{\noexpand\cellcolor{ColorMid!\Lightness!white}}\temp        \else
                \pgfmathparse{((#1 - 10) / 15) * 100}
    \let\Lightness\pgfmathresult
            \edef\temp{\noexpand\cellcolor{ColorLow!\Lightness!white}}\temp % Lower values get lighter color
        \fi
    \fi
    #1
}

\newcommand{\applyGradienth}[1]{
    \ifdim #1 pt > 35 pt
    \pgfmathparse{((#1 - 35) / 5) * 100}
    \let\Lightness\pgfmathresult
    \edef\temp{\noexpand\cellcolor{ColorHigh!\Lightness!white}}\temp
    \else
        \ifdim #1 pt > 30 pt
            \pgfmathparse{((#1 - 30) / 5) * 100}
                \let\Lightness\pgfmathresult
                \edef\temp{\noexpand\cellcolor{ColorMid!\Lightness!white}}\temp        \else
                \pgfmathparse{((#1 - 25) / 5) * 100}
    \let\Lightness\pgfmathresult
            \edef\temp{\noexpand\cellcolor{ColorLow!\Lightness!white}}\temp % Lower values get lighter color
        \fi
    \fi
    #1
}

\newcommand{\applyGradienti}[1]{
    \ifdim #1 pt > 44 pt
    \pgfmathparse{((#1 - 44) / 4) * 100}
    \let\Lightness\pgfmathresult
    \edef\temp{\noexpand\cellcolor{ColorHigh!\Lightness!white}}\temp
    \else
        \ifdim #1 pt > 42 pt
            \pgfmathparse{((#1 - 42) / 2) * 100}
                \let\Lightness\pgfmathresult
                \edef\temp{\noexpand\cellcolor{ColorMid!\Lightness!white}}\temp     
                \else
                \pgfmathparse{((#1 - 40) / 2) * 100}
    \let\Lightness\pgfmathresult
            \edef\temp{\noexpand\cellcolor{ColorLow!\Lightness!white}}\temp % Lower values get lighter color
        \fi
    \fi
    #1
}

\newcommand{\applyGradientj}[1]{
    \ifdim #1 pt > 44 pt
    \pgfmathparse{((#1 - 44) / 4) * 100}
    \let\Lightness\pgfmathresult
    \edef\temp{\noexpand\cellcolor{ColorHigh!\Lightness!white}}\temp
    \else
        \ifdim #1 pt > 42 pt
            \pgfmathparse{((#1 - 42) / 2) * 100}
                \let\Lightness\pgfmathresult
                \edef\temp{\noexpand\cellcolor{ColorMid!\Lightness!white}}\temp     
                \else
                \pgfmathparse{((#1 - 40) / 2) * 100}
    \let\Lightness\pgfmathresult
            \edef\temp{\noexpand\cellcolor{ColorLow!\Lightness!white}}\temp % Lower values get lighter color
        \fi
    \fi
    #1
}

\title{Tables as Texts or Images: \\Evaluating the Table Reasoning Ability of LLMs and MLLMs}

\author{Naihao Deng*$^{\twemoji{peach}}$, Zhenjie Sun*$^{\twemoji{peach}}$, Ruiqi He$^{\twemoji{peach}}$, 
Aman Sikka$^{\twemoji{peach}}$, \\
{\bf Yulong Chen$^{\twemoji{blueberries}}$, Lin Ma$^{\twemoji{peach}}$, Yue Zhang$^{\twemoji{lemon}}$, Rada Mihalcea$^{\twemoji{peach}}$}\\
$^{\twemoji{peach}}$ University of Michigan\quad$^{\twemoji{blueberries}}$ University of Cambridge\quad$^{\twemoji{lemon}}$ Westlake University\\
  \texttt{\{dnaihao, zjsun\}@umich.edu} \\}

\begin{document}
\maketitle

\def\thefootnote{*}\footnotetext{Contributed equally to this work.
\Cref{app-sec: contributions} lists the detailed contributions.}

\begin{abstract}
%Tables contrast with unstructured text data by its structure to organize the information.
In this paper, we investigate the effectiveness of various LLMs in interpreting tabular data through different prompting strategies and data formats. 
Our analyses extend across six benchmarks for table-related tasks such as question-answering and fact-checking. 
We introduce for the first time the assessment of LLMs' performance on image-based table representations. 
Specifically, we compare five text-based and three image-based table representations, demonstrating the role of representation and prompting on LLM performance. 
Our study provides insights into the effective use of LLMs on table-related tasks.
Our data is available at: \url{https://github.com/dnaihao/Tables-as-Texts-or-Images}.

\end{abstract}

\section{Introduction}

\begin{figure*}[t]
\centering
\includegraphics[width=0.98\linewidth]{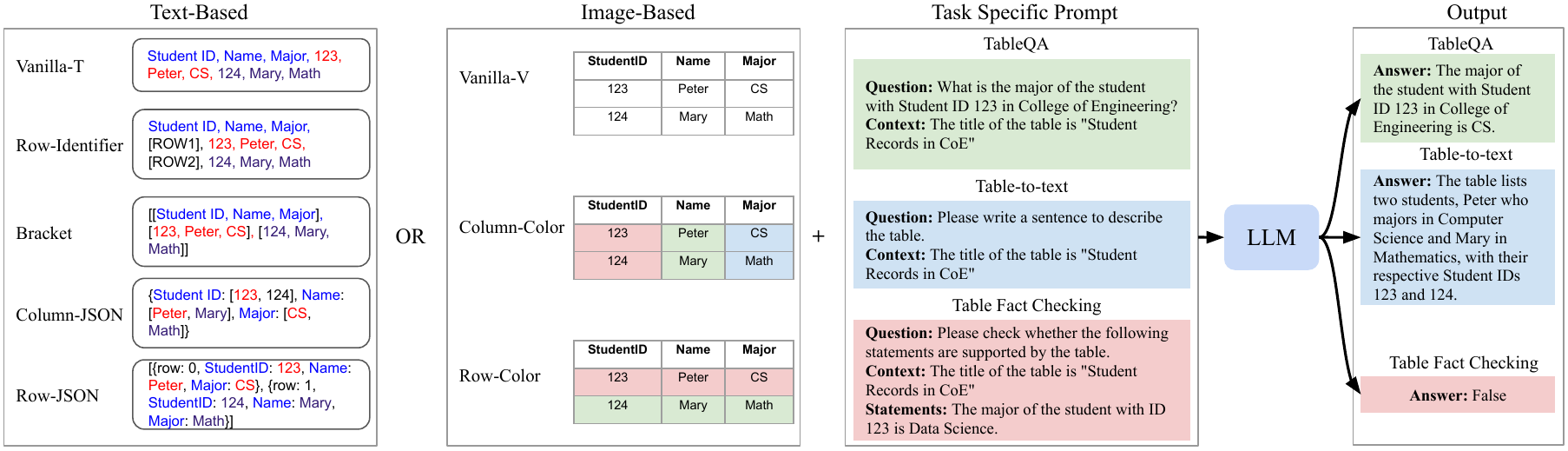} 
\caption{Concept diagram.
In this paper, we study differences in table representations.
For each example, we prompt LLMs with the question and the context information, as well as one of the table representations.}
\label{fig:concept-diagram}
\end{figure*}

Recent years have witnessed an explosion of Large Language Models (LLMs), with impressive performance on various Natural Language Processing (NLP) tasks \cite{brown2020language, touvron2023Llama, team2023gemini}.
Research to date has examined the performance of LLMs for various aspects and abilities \cite{bang2023multitask, bubeck2023sparks, akter2023depth}, but their effectiveness on  structured data such as tables is less explored.

Unlike unstructured text, tables are systematically organized structures of a large amount of information. 
This characteristic makes tabular data serve as the foundations for numerous applications, including medical diagnostics, virtual personal assistants, customer relationship management \cite{hemphill-etal-1990-atis, dahl-etal-1994-expanding, akhtar-etal-2022-pubhealthtab, xie-etal-2022-unifiedskg}, etc.

The evaluation of LLMs on processing tabular data involves many challenges.
First, there are many ways to represent the information in tables.
If we represent the table in pure text, we may use naive linearization or insert brackets to better represent table structures. 
Meanwhile, emerging multimodal LLMs like GPT-4 \cite{achiam2023gpt} and Gemini \cite{team2023gemini} offer image-based approaches, where we can pass the table as images to the LLMs.
In such cases, visual cues like color highlighting in tables can influence outcomes.
Second, diverse prompting methods for text may also apply to tabular data, which can yield varied results \cite{wei2022chain}.
Furthermore, the tasks involving tabular data are diverse, including table fact-checking \cite{chen2019tabfact} and table question answering \cite{pasupat-liang-2015-compositional}, and table-to-text generation \cite{novikova-etal-2017-e2e}, etc.

In this paper, we systematically evaluate model performance on tabular data for both textual LLMs and multi-modal LLMs.
Specifically, we investigate several research questions, including the effectiveness of image-based representation of tabular data and how different text-based or image-based prompt methods affect LLMs' performance on table-related tasks. 
In addition, we provide analysis and hypothesis of LLMs’ behaviors.
% We investigate three prompting strategies, including vanilla prompting, chain-of-thought prompting \cite{wei2022chain} and expert prompting \cite{xu2023expertprompting}.
% For tabular data representation, we explore five text-based formats, including (1) direct linearization, (2) adding a row identifier (3) using brackets to separate rows (4) JSON format with column names the key (5) JSON format with each row as a JSON object,
% alongside three image-based approaches, including (1) plain table image without colors (2) colored table image to distinguish rows (3) colored table image to distinguish columns.
% Our study encompasses six existing benchmarks covering tasks of table question answering, fact-checking, and text generation.
% Furthermore, we conduct our studies on a wide range of LLMs, including closed-source models such as Gemini, GPT-3.5, and GPT-4, as well as open-source alternatives such as Llama-2.
Our findings include:
\begin{itemize}[leftmargin=0.4cm,itemsep=0.1cm]
    \item LLMs maintain decent performance when we use image-based table representations. 
    Sometimes, image-based table representations can make LLMs perform better.
    \item There are nuances in the prompting design for table-related tasks, revealed by our comparisons of various prompting methods for text- and image-based table representations.
\end{itemize}

To the best of our knowledge, we are the first to study how LLMs perform with image-based table representations. We believe this paper draws new insights into optimizing table-based information processing.

\section{Related Work}

\paragraph{Table-Related Tasks.} 
Tasks involving structured data have attracted interest in various tasks from diverse communities \cite{deng2020turl, chen-etal-2021-hitter, chen-etal-2022-towards-table, deng-etal-2022-recent}, among which there is a huge focus on tabular data \cite{yin-etal-2020-tabert, herzig-etal-2020-tapas}. 

Researchers have investigated various ways to encode tabular data. 
\citet{hwang2019comprehensive, liu2021tapex, cong2024observatory} linearize the table content.
Others employ model-specific techniques such as adapting the attention mechanism to better align transformer-based models with the tabular data \cite{zhang-etal-2020-table, yang-etal-2022-tableformer} or designing hierarchical encoding to capture the table structure \cite{wang2021tuta}, further tuning LLMs on tabular data \cite{zha2023tablegpt, zhang2023tableLlama}, etc. 
In contrast, our work focuses on exploring various table representations and prompts LLMs directly.

\paragraph{Prompting LLMs.} 
Researchers have prompted LLMs to evaluate LLMs' performance on traditional NLP tasks \cite{bang-etal-2023-multitask} as well as on various complex reasoning tasks \cite{NEURIPS2022_b654d615, wu-etal-2023-hi, zheng2024gpt}.
On the contrary, to the best of our knowledge, few works have prompted these LLMs on tasks involving tabular data.

For closed-source LLMs, researchers adopt hard prompts to manually craft text prompts with discrete tokens \cite{qiao2022reasoning, bahng2022exploring, liu2023pre}.
\citet{wei2022chain} develop chain-of-though prompting, \citet{xu2023expertprompting} develop expert prompting.
In our work, we include the comparison between vanilla, chain-of-thought, and expert prompting for LLMs on table-related tasks.

\section{Experiment Setups}

\subsection{Experimented LLMs}
\label{subsec: experimented-llms}

\begin{table}[t]
\centering
\renewcommand*{\arraystretch}{1.1}
\small
\begin{tabular}{lcccc}
\toprule
 \textbf{Models} & \textbf{\# P(B)} & \textbf{\opensource \,/\closedsource} & \textbf{\makecell{+V?}} & \textbf{Company} \\
\midrule
\textbf{Llama-2} & 7/13/70 & \opensource & \usym{2715} & Meta \\
% \textbf{Bloom} & 176 & \faUnlockAlt & \usym{2715} & Huggingface \\
\textbf{GPT-3.5} & -- & \closedsource & \usym{2715} & OpenAI \\
\textbf{GPT-4} & -- & \closedsource & \usym{1F5F8} & OpenAI \\
\textbf{Gemini}$_\text{pro}$ & -- & \closedsource & \usym{1F5F8} & Google \\
\midrule
\textbf{Llama-3} & 8/70 & \opensource & \usym{2715} & Meta\\
\textbf{Gemma} & 2/7 & \opensource & \usym{1F5F8} & Meta \\
\textbf{GPT-4o} & - & \closedsource &  \usym{1F5F8} & OpenAI \\
\bottomrule
\end{tabular}
\caption{Comparison of LLMs used in our experiments. 
``\# P'' represents the number of parameters in billions (B).
Note that we do not include the number of parameters for the closed-source models as there are no official documents revealing this information. 
``\opensource \,/\closedsource'' indicates whether the LLM is open-source (\opensource) or closed-source (\closedsource).
``+V?'' indicates whether the visual input is allowed for the LLM.
``Company'' indicates which company the LLM is from.
We conduct the majority of our experiments with Llama-2. GPT-3.5, GPT-4 and Gemini$_\text{pro}$.
We prompt the January 2024 versions of the GPT-3.5 and GPT-4.
We include additional experiments for GPT-4o, Llama-3, and Gemma models in \Cref{app-sec: addition-experiments}.
}
\label{table:llms-comparison}
\end{table}

\Cref{table:llms-comparison} describes the LLMs we use for our experiments.
We use closed-source models such as GPT-3.5 and GPT-4 \cite{brown2020language, ouyang2022training}, and Gemini \cite{team2023gemini}. 
We note that GPT-4 and Gemini are multimodal models, which can take tables as images.
For open-source models, we use 
the chat models from Llama-2 \cite{touvron2023Llama} families from the 7 billion to the 70 billion parameter version as they are claimed to perform on par with closed-source models like ChatGPT.\footnote{https://huggingface.co/meta-Llama/Llama-2-70b-chat}

\begin{table}[t]
    \centering
    \includegraphics[width=0.95\linewidth]{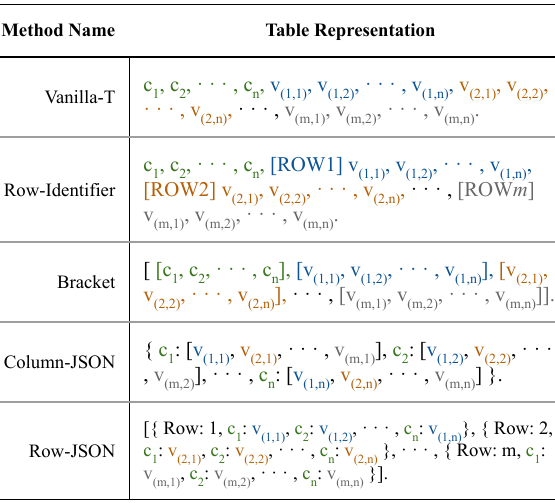}
    \caption{Text-based table representation examples.
    We construct the examples assuming a table of $m$ rows and $n$ columns, where c$_i$ denotes the column name of column $i$ and v$_{(i, j)}$ denotes the cell value at row $i$ and column $j$.
    We use colored text to indicate different rows in the table to assist readers.}
    \label{tab:text-based-table-representation}
\end{table}

\subsection{Prompting Strategies}
\label{subsec: prompting-strategies}
We explore two ways to represent tables in the prompt, \textbf{Text-Based} and \textbf{Image-Based}.

\paragraph{Text-Based.}
Apart from the information contained in the cells of tables, the structure of the table maintains information such as what cell values are in the same row or column, and what cell values correspond to a particular column.
Therefore, we explore various ways to incorporate such structure information into the text prompt.

\begin{table*}[t]
    \small
    \centering
    \begin{tabular}{cccccc}
        \toprule
        Task Family & Name & Domain & Input & Output & Metrics\\
        \midrule
        \multirow{2}{*}{Table QA}&WikiTQ \cite{pasupat-liang-2015-compositional} & Wikipedia & Table & Text & Acc\\
        &FinQA \cite{chen-etal-2021-finqa} & Finance & Table + Text & Text & Acc \\
        \midrule
        Table Fact Checking & TabFact \cite{chen2019tabfact} & Wikipedia & Table & Boolean & Acc \\
        \midrule
        \multirow{3}{*}{Table-to-text} & E2E \cite{novikova-etal-2017-e2e} & Restaurants & Table & Text & ROUGE, Human\\
        & ToTTo \cite{parikh-etal-2020-totto} & Wikipedia & Table + Text & Text & ROUGE\\
        & LogicNLG \cite{chen-etal-2020-logical} & Wikipedia & Table + Text & Entity & Acc \\
        \bottomrule
    \end{tabular}
    \caption{Dataset descriptions.
    For Input, we refer to the input information other than the question, the statement for fact-checking, or the statement that requires the model to describe the table content.}
    \label{tab:dataset-description}
\end{table*}

\begin{itemize}[leftmargin=0.4cm]
    \item \textbf{Vanilla-T} 
lists column names followed by cell values in each row sequentially, an approach adopted in various prior works \cite{hwang2019comprehensive, liu2021tapex}.
    \item \textbf{Row-Identifier} 
adds an identifier as the prefix for each row to distinguish different rows in the linearized table sequence.
    \item \textbf{Bracket} 
encloses the column names and their values in brackets to distinguish each row.
    \item \textbf{Column-JSON} 
represents the table in JSON format, where column names are the keys that map to the list of cell values corresponding to that column.
    \item \textbf{Row-JSON}
represents each row as a JSON object, within which the column names and their corresponding cell values are represented as key-value pairs.
\end{itemize}

\Cref{tab:text-based-table-representation} shows examples of these text-based table representations.

\begin{figure}[t]
    \centering
    % First row of subfigures
\includegraphics[width=1\linewidth]{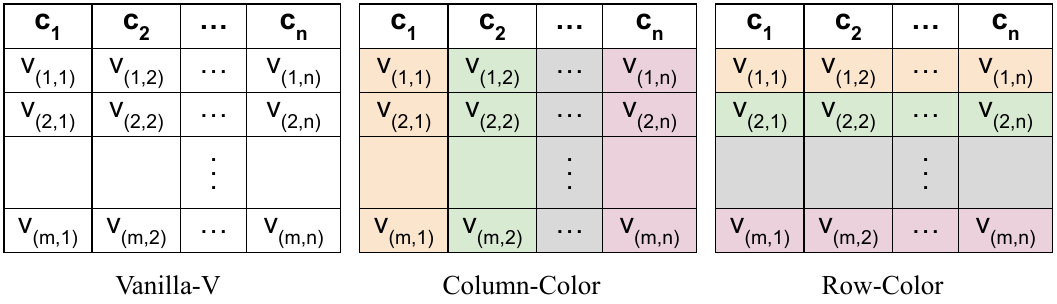} 
    \caption{Image-based table representation examples.
    We construct these examples based on the same table described in \Cref{tab:text-based-table-representation}.}
    \label{fig:image-based-example}
\end{figure}

\paragraph{Image-Based.}
Alternatively, we can pass the table as an image to the recent multimodal LLMs such as GPT-4 and Gemini.
In this way, LLMs would ``view'' the table in a similar way as how we human beings view the table.
We explore various table-highlighting methods as different visual cues may influence the model outcomes as shown by \citet{shtedritski2023does} who study how highlighting can influence CLIP model \cite{radford2021learning}'s performance on vision and language tasks.
We pass these images of the table to LLMs.

\begin{figure*}[t]
\centering
\includegraphics[width=0.95\linewidth]{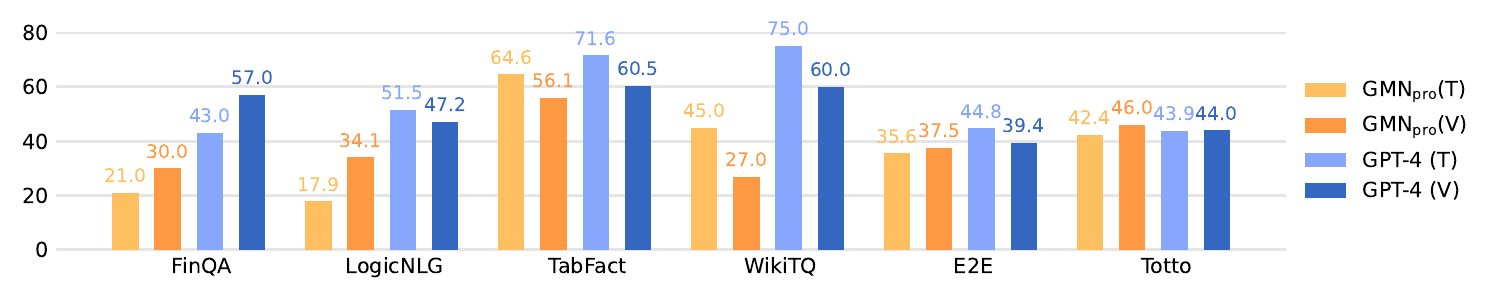} 
\caption{Performance comparison between passing the text versus image representations of tables to GPT-4 and Gemini$_\text{Pro}$ across FinQA, LogicNLG, TabFact, and WikiTQ by accuracy, and E2E and ToTTo by ROUGE-L scores.
We feed the linearized table (Vanilla-T) as the text-based representation, and the original table image (Vanilla-V) as the image-based representation to these LLMs.}
\label{fig:text_vision_comparison}
\end{figure*}

\begin{itemize}[leftmargin=0.4cm]
    \item \textbf{Vanilla-V}
feeds the table image without any colors or highlighting to LLMs.
    \item \textbf{Column-Color}
uses a single color for each table column.
Therefore, the LLM may easily distinguish columns as cells in the same column are annotated by the same color, whereas different colors annotate cells from different columns.
    \item \textbf{Row-Color}
uses a single color for each row in the table.
The same color annotates cells in the same row, whereas different colors annotate cells in different rows.
\end{itemize}

\Cref{fig:image-based-example} show examples for these image-based table representations.

On top of different methods to represent tables, we test the vanilla prompting, chain-of-thought prompting \cite{wei2022chain}, and expert prompting \cite{xu2023expertprompting} by adding ``let's pretend you are an expert in reading and understanding tables'' to the prompt.
\Cref{app-sec: prompt-examples} provides an example for each table representation and prompting method.

\subsection{Datasets}

We make use of six previously introduced datasets that cover different table sources such as Wikipedia and financial reports, examine model abilities such as information extraction and arithmetic reasoning, and cover table-related tasks such as table question answering, table fact-checking, and table-to-text generation.
\Cref{tab:dataset-description} provides information for each dataset we use.
Considering the limited access to LLMs' APIs and the scale of the comparison, we randomly select 100 examples from the test set for each of these datasets to conduct our analysis. 
In total, we run the our experiments on 54,000 instances\footnote{5 text form table representations $\times$ 3 prompting strategies $\times$ 6 LLMs $\times$ 6 datasets $\times$ 100 examples / dataset + 3 image table representations $\times$ 3 $\times$ prompting strategies $\times$ 2 MLLMs $\times$ 6 datasets $\times$ 100 examples = 54,000 instances. 
This does not include the number of additional experiments we run on other models such as Gemma, Llama-3 and GPT-4o.}.
In addition, we have observed significant differences between LLMs’ performance and report the results from three significance tests in \Cref{app-sec:significance-test}.

\subsection{Metrics}
\label{subsec: metrics}
% We use both automatic and human evaluations.
Following \citet{pasupat-liang-2015-compositional, chen2019tabfact, chen-etal-2020-logical, chen-etal-2021-finqa}, we compute accuracy scores on WikiTQ, TabFact, LogicNLG, FinQA.

We adopt the automatic ROUGE evaluation for table-to-text generation datasets ToTTo and E2E. 
In addition, the authors manually investigate the generation quality on the E2E dataset by whether the generation encapsulates the table information without any additional information that cannot be inferred from the table.

\section{Research Questions}

Using the setup described previously, we can now seek answers to several research questions concerning the use of LLMs for tabular data.

\subsection*{RQ1. Are image-based representations of tabular data effective?}

\paragraph{Test:} We compare using the linearized table representation (Vanilla-T in text-based table representation) and the table image (Vanilla-V in image-based table representation) as the input for both GPT-4 model and Gemini$_\text{pro}$.
We use vanilla prompting in this comparison and plot \Cref{fig:text_vision_comparison}.
We report the results for other prompting methods in \Cref{fig:text_vision_comparison_cot,fig:text_vision_comparison_expert} in \Cref{app-subsec:rq1-contd}.

\paragraph{TL;DR Answer: Yes.}

\begin{figure}[t]
    \centering
    \includegraphics[width=0.95\linewidth]{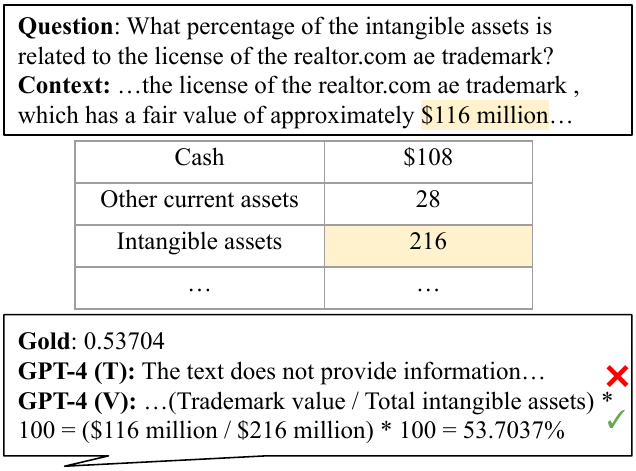}
    \caption{An example from FinQA. 
    We highlight the relevant parts from the context and the table and omit irrelevant parts to help readers.
    We feed the linearized table (Vanilla-T) as the text-based representation (GPT-4 (T)), and the original table image (Vanilla-V) as the image-based representation to GPT-4 (GPT-4 (V)).}
    \label{fig:finqa-rq1-example}
\end{figure}

\paragraph{Full Answer:}
\Cref{fig:text_vision_comparison} shows that in most cases, LLMs perform comparably if we represent tables as images versus text.
On datasets such as FinQA, passing image representation of tables to Gemini$_\text{pro}$ and GPT-4 outperform passing text representations of the tables significantly.
As FinQA focuses on financial question answering with long context and many numerical relations, we hypothesize that \textit{representing tables as images can help LLMs in complex reasoning}. 
Since these multimodal LLMs have a strong capability over visual input \cite{yang2023dawn}, representing tables as images may reduce the cognitive load for LLMs to parse and understand dense text.
This is especially beneficial when the context involves long passages of text that may also contain numerous numerical relations.
As shown in \Cref{fig:finqa-rq1-example}, since the context is long (around 416 English words, approximately 556 tokens for GPT models) and involves various numerical relations, GPT-4 ignores the relevant clues in text when we pass text representation of the table. 

In contrast, when we pass the table image, GPT-4 can effectively leverage information from both the text and visual modality for its reasoning process.

On WikiTQ and TabFact, both Gemini$_\text{pro}$ and GPT-4 perform better with the text than the image representation of the table significantly.
We notice that both datasets are sourced from Wikipedia and the texts from Wikipedia are commonly used to pre-train LLMs \cite{brown2020language, touvron2023Llama}.
GPT-4 and Gemini$_\text{pro}$ may have encountered these tables in their pre-training phase in the text format rather than the image format, leading to the performance disparity between text and image representation of tables for both Gemini$_\text{pro}$ and GPT-4 on WikiTQ and TabFact.

\subsection*{RQ2. How do different text-based prompt methods affect LLMs' performance on table-related tasks?}
\paragraph{Test:} We compare the five text-based table representations introduced in \Cref{subsec: prompting-strategies}.
On top of the five representations, we also compare how vanilla, chain-of-thought, and expert prompting affect the model performance.
We conduct the comparison using all six LLMs in \Cref{subsec: experimented-llms} and average their accuracy scores across FinQA, LogicNLG, TabFact, and WikiTQ.
\Cref{app-subsec:rq2-contd} reports LLMs' performance on E2E and ToTTo datasets.

\begin{table}[t]
    \small
    \centering
    \setlength{\tabcolsep}{2.5pt}
    \begin{tabular}{ccccccc}
        \toprule
        & \multicolumn{2}{c}{GPT} & \multirow{2}{*}{GMN$_\text{pro}$} & \multicolumn{3}{c}{Llama-2}\\
        &3.5 & 4& & 7B & 13B & 70B \\
        \midrule
        \multicolumn{2}{l}{\textbf{Vanilla-T}} & & & & & \\
        V & \applyGradient{52.5} & \applyGradient{60.3} & \applyGradient{37.1} & \applyGradient{28.8} & \applyGradient{35.3} & \applyGradient{42.7} \\
E & \applyGradient{51.0} & \textbf{\applyGradient{63.8}} & \applyGradient{39.5} & \applyGradient{29.0} & \applyGradient{35.1} & \applyGradient{46.7} \\
 CoT & \applyGradient{55.2} & \applyGradient{62.6} & \applyGradient{53.5} & \applyGradient{32.1} & \applyGradient{37.6} & \applyGradient{48.3} \\
\midrule
\textbf{Brackt} & & & & & & \\
V & \applyGradient{50.9} & \applyGradient{60.1} & \applyGradient{38.4} & \applyGradient{28.4} & \applyGradient{36.6} & \applyGradient{42.2} \\
E & \applyGradient{47.9} & \applyGradient{62.8} & \applyGradient{39.5} & \applyGradient{28.1} & \applyGradient{34.5} & \applyGradient{45.8} \\
 CoT & \applyGradient{51.4} & \applyGradient{61.9} & \applyGradient{57.3} & \applyGradient{34.2} & \applyGradient{39.3} & \applyGradient{50.0} \\
        \bottomrule
    \end{tabular}
    \caption{For text-based table representations, averaged accuracy scores across FinQA, LogicNLG, TabFact, and WikiTQ for different LLMs.
    ``GMN$_\text{pro}$'' represents Gemini$_\text{pro}$ model, ``V'', ``E'', and ``CoT'' represent vanilla, expert and chain-of-thought prompting, respectively.}
    \label{tab:rq2-vanilla-bracket}
\end{table}

\paragraph{TL;DR Answer 2.1:} \textbf{Expert prompting works the best when the LLM is an ``expert''.}

\paragraph{Full Answer 2.1:} 
With respect to vanilla, CoT, and expert prompting, for GPT-4, we note that expert prompting outperforms the other two prompting methods consistently.
For instance, for the vanilla linearized table representations (Vanilla-T), expert prompting outperforms the CoT and the vanilla prompting method by 1.2\% and 3.5\%, respectively (\Cref{tab:rq2-vanilla-bracket}).
In contrast, CoT prompting instead of expert prompting leads to the best performance for all other models.
For instance, for GPT-3.5 with Vanilla-T table representation, CoT prompting outperforms vanilla and expert prompting by 2.7\% and 4.2\% (\Cref{tab:rq2-vanilla-bracket}).

On the other hand, GPT-4 outperforms all other models, as the best average score GPT-4 achieves is 63.8\%, compared to 55.2\% by GPT-3.5 and 50.0\% by Llama-2-70B.
We suspect that because of GPT-4's ``expertise'' on these tasks, expert prompting can further enhance its reasoning ability as GPT-4 can ``pretend they are an expert in reading and understanding tables''.
In contrast, expert prompting may not fit the less capable LLMs as they may not ``pretend an expert'' well.

\paragraph{TL;DR Answer 2.2: CoT prompting can sometimes boost up the performance significantly.} 

\paragraph{Full Answer 2.2:} We notice that CoT prompting significantly improves Gemini$_\text{pro}$'s performance from 38.4\% to 57.3\% using the bracket table representation (\Cref{tab:rq2-vanilla-bracket}), which outperforms the best performance 55.2\% by GPT-3.5. 
This suggests that proper prompting can make a big difference in LLMs' performance and unleash the potential within the LLM.
On the other hand, it underscores the complexity of LLMs' evaluation and the importance of prompt engineering, as we may underestimate an LLMs' performance because of an improper prompt.

\paragraph{TL;DR Answer 2.3: Bracket representation can help LLMs better understand tables.}

\paragraph{Full Answer 2.3:}
Compared to linearizing tables directly (Vanilla-T), adding brackets to distinguish rows in the table boosts up model performances for Gemini$_\text{pro}$ and different versions of Llama-2 models (\Cref{tab:rq2-vanilla-bracket}). 

\begin{figure}[t]
    \centering
    \includegraphics[width=0.95\linewidth]{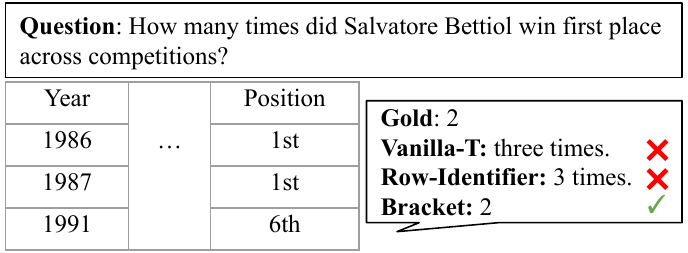}
    \caption{An example from WikiTQ. 
    We use Gemini$_\text{pro}$ with vanilla prompting and show its prediction when we use the linearized table representation (Vanilla-T), insert ``Row-Identifier'' or ``Bracket'' in the representation.}
    \label{fig:gemini-pro-bracket-vs-plain-rq2}
\end{figure}

\Cref{fig:gemini-pro-bracket-vs-plain-rq2} shows a WikiTQ example from Gemini$_\text{pro}$, where the vanilla prompting fails to count the number of ``1st'' place.
We suspect that the simple linearized table representation does not have a clear boundary between rows, which may lead to confusion or misinterpretation of data relationships.
In addition, adding the row identifier in the sequence does not help while the LLM answers correctly with the bracket representation.
We conjecture that LLMs may be familiar with brackets from their pre-training exposure.
Since brackets are fundamental components of many programming languages, and Github which contains rich code is often used as a source for pre-training corpora \cite{touvron2023Llama}, LLMs may have acquired proficiency in recognizing and interpreting bracketed structures.

\paragraph{TL;DR Answer 2.4: Different table representations do not affect the performance of GPT models much.}
\paragraph{Full Answer 2.4:}
% On the contrary, for GPT-3.5 and GPT-4 models, linearizing tables directly (Vanilla-T) obtains the best performance.
Even without any sophisticated prompting methods, the GPT-3.5 and GPT-4 achieve a decent performance (52.5\% and 60.3\% respectively using the vanilla prompting and linearized table representation from \Cref{tab:rq2-vanilla-bracket}), demonstrating their strong table understanding abilities.
In such cases, brackets or other kinds of table representations may add extra ``workload'' to the model, which dilutes the models' attention to the original table content and thus leads to worse performance.

\begin{table}[t]
    \small
    \centering
    \begin{tabular}{ccccccc}
    \toprule
         &  \multicolumn{3}{c}{GPT-4} & \multicolumn{3}{c}{Gemini$_\text{pro}$}\\
         & V  & E & CoT & V & E & CoT \\
    \midrule
    
    VV & \applyGradienta{56.2} & \applyGradienta{54.9} & \applyGradienta{57.8} & \applyGradienta{36.8} & \applyGradienta{37.2} & \applyGradienta{46.0} \\
CC & \applyGradienta{53.3} & \applyGradienta{52.8} & \applyGradienta{58.0} & \applyGradienta{37.1} & \applyGradienta{37.8} & \applyGradienta{45.1} \\
RC & \applyGradienta{51.8} & \applyGradienta{51.6} & \textbf{\applyGradienta{60.2}} & \applyGradienta{39.4} & \applyGradienta{38.7} & \applyGradienta{46.2} \\
    
    \bottomrule 
    \end{tabular}
    \caption{For image-based table representations, averaged accuracy scores across FinQA, LogicNLG, TabFact, and WikiTQ for GPT-4 and Gemini$_\text{pro}$.
    For the headers, ``V'', ``E'', and ''CoT'' represent vanilla, expert, and chain-of-thought prompting, respectively.
    For the row names, ``VV'', ``CC', and ``RC'' represent Vanilla-V, Column-Color, and Row-Color, respectively.}
    \label{tab:image-prompting-rq3}
\end{table}

% \begin{table}[t]
%     \small
%     \centering
%     \begin{tabular}{ccccccc}
%     \toprule
%          &  \multicolumn{2}{c}{GPT-4} & \multicolumn{2}{c}{Gemini$_\text{pro}$}\\
%          & V  & CoT & V  & CoT \\
%     \midrule
    
%     VV & \applyGradienta{56.2}  & \applyGradienta{57.8} & \applyGradienta{36.8}  & \applyGradienta{46.0} \\
% CC & \applyGradienta{53.3} & \applyGradienta{58.0} & \applyGradienta{37.1}  & \applyGradienta{45.1} \\
% RC & \applyGradienta{51.8} & \textbf{\applyGradienta{60.2}} & \applyGradienta{39.4}  & \applyGradienta{46.2} \\
    
%     \bottomrule 
%     \end{tabular}
%     \caption{For image-based table representations, averaged accuracy scores across FinQA, LogicNLG, TabFact, and WikiTQ for GPT-4 and Gemini$_\text{pro}$.
%     For the headers, ``V'', ``E'', and ''CoT'' represent vanilla, expert, and chain-of-thought prompting, respectively.
%     For the row names, ``VV'', ``CC', and ``RC'' represent Vanilla-V, Column-Color, and Row-Color, respectively.}
%     \label{tab:image-prompting-rq3}
% \end{table}

\begin{figure}[t]
    \centering
    \includegraphics[width=0.95\linewidth]{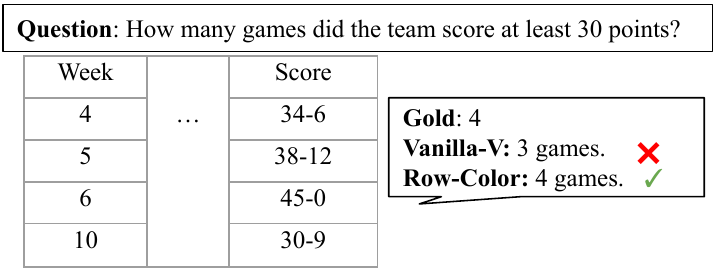}
    \caption{An example from WikiTQ. 
    We use Gemini$_\text{pro}$ with vanilla prompting and show its prediction when we use the original table image (Vanilla-V) and the table image that uses different colors to distinguish rows in the table (Row-Color).}
    \label{fig:gemini-pro-bracket-vs-plain-rq3}
\end{figure}

\subsection*{RQ3. How do different image-based prompt methods affect LLMs' performance on table-related tasks?}

\paragraph{Test:}
We test the three image-based table representations in \Cref{subsec: prompting-strategies} together with vanilla, chain-of-thought, and expert prompting.
We test the Gemini$_\text{pro}$ and GPT-4 model which can take images as the input.
We average the accuracy scores across FinQA, LogicNLG, TabFact, and WikiTQ.
\Cref{app-subsec:rq3-contd} reports LLMs' performance on E2E and ToTTo datasets.

\paragraph{TL;DR Answer 3.1: CoT prompting helps LLMs reason over images of the table.}

\paragraph{Full Answer 3.1:}
In \Cref{tab:image-prompting-rq3}, we observe that chain-of-thought prompting helps multimodal LLMs in all image-based table representations.
For instance, when using different colors to distinguish rows in the table (Row-Color), the average accuracy score for GPT-4 improves from 51.8\% by vanilla prompting to 60.2\% by chain-of-thought prompting.
By explicitly outlining the reasoning process, chain-of-thought prompting may help LLMs better understand the context and relationships between different rows and columns in the table, therefore better aligning this visual information with the question text. 
Such consistent performance improvements suggest that chain-of-thought prompting may enhance information fusion across the text and vision modality.

\paragraph{TL;DR Answer 3.2: Distinguishing rows may lead to better performance for LLMs to reason over images of the table.}

\paragraph{Full Answer 3.2:}
In \Cref{tab:image-prompting-rq3}, under CoT prompting, GPT4 performs slightly better when using colors to distinguish different rows, which also yields the overall best performance using images of the table.
In contrast, under CoT prompting, using colors to distinguish columns yields similar performance to vanilla image (58.0\% to 57.8\% for GPT-4 and 45.1\% to 46.0\% for Gemini$_\text{pro}$), suggesting that these advanced LLMs may not capture row information as well as column information.

\Cref{fig:gemini-pro-bracket-vs-plain-rq3} shows a WikiTQ example with Gemini$_\text{pro}$ model's predictions.
Since the question asks about the number of games, it requires the model to count how many rows satisfy such a condition.
Using colors to distinguish rows may help models visually segment and categorize the data.
This visual differentiation may act as a cognitive aid, which reduces the complexity of parsing and interpreting the tabular data.

\paragraph{TL;DR Answer 3.3: The more capable LLM does not necessarily benefit more from the colored images.}

\paragraph{Full Answer 3.3:} 
In addition, if we use the vanilla prompt, the different coloring methods may even hurt the performance of GPT-4 (for GPT-4, coloring rows with different colors yields 51.8\% compared to 56.2\% without adding any color), but helpful for Gemini$_\text{pro}$ (for Gemini$_\text{pro}$, coloring rows with different colors yields 39.4\% compared to 36.8\% without adding any color). 
This suggests that the effectiveness of how different LLMs can leverage colored images varies, and does not depend on the model's overall performance.

\begin{table}[t]
    \small
    \centering
    \setlength{\tabcolsep}{2.3pt}
    \begin{tabular}{cc|cccccc}
        \toprule
        \multirow{2}{*}{Rep} & \multirow{2}{*}{Cues} & \multicolumn{2}{c}{GPT} & \multirow{2}{*}{GMN$_\text{pro}$} & \multicolumn{3}{c}{Llama-2}\\
        & &3.5 & 4& & 7B & 13B & 70B \\
        \midrule
        T & N/A & \applyGradientb{34} & \applyGradientb{43} & \applyGradientb{21} & \applyGradientb{10} & \applyGradientb{20} & \applyGradientb{41} \\
        T & T & \applyGradientb{30} & \applyGradientb{51} & \applyGradientb{25} & \applyGradientb{14} & \applyGradientb{16} & \applyGradientb{37} \\
        V & N/A & - & \applyGradientb{57} & \applyGradientb{30} & - & - & -\\
        V & T & - & \applyGradientb{58} & \applyGradientb{34} & - & - & -\\
        V & V & - & \applyGradientb{57} & \applyGradientb{28} & - & - & -\\
        V & V+T & - & \textbf{\applyGradientb{61}} & \applyGradientb{38} & - & - & -\\
        \bottomrule
    \end{tabular}
    \caption{Accuracy scores of LLMs on FinQA. 
    We use vanilla prompting across experiments in this table.
    GMN$_\text{pro}$ represents Gemini$_\text{pro}$ model.
    We denote text and image-based table representations as ``T'' and ``V'' in the ``Rep'' column, respectively.
    The ``Cues'' column indicates how we highlight the relevant cells, where ``N/A'' indicates no information about relevant cells, ``T'' indicates referring to relevant cells in the text, ``V'' indicates highlighting relevant cells on the table image, ``V + T'' indicates both highlighting relevant cells on the table image and referring to them in the text.
    }
    \label{tab: highlight-vs-non-highlight}
\end{table}

\subsection*{RQ4. Does highlighting relevant cells yield a better performance?}

\paragraph{Test:} We test all six LLMs in \Cref{subsec: experimented-llms} on FinQA which provides relevant cells in the table for each instance.
We refer to the relevant cells by adding ``Please pay attention to the highlighted cells: (row index, column index, cell value)'' in the text prompt, or mark them on the table image directly.
\Cref{app-sec: prompt-examples} provides our prompt examples.
We use vanilla prompting in this comparison.

\paragraph{TL;DR Answer: Yes.}
\paragraph{Full Answer:}
In \Cref{tab: highlight-vs-non-highlight}, we notice that in most cases, referring LLMs to specific cells helps LLMs better attend to them, thereby helping LLMs reason over the example. 
However, LLMs' performance may get hurt when we refer to the relevant cells through text such as Llama-2-13B and 70B.
This may be due to the inherent limitations of textual descriptions for conveying spatial or relational information.
In order to relate the mentioned cells in the text, the model needs to figure out the connection between the mentioned cell and the cell in the linearized table, which can be challenging to the model given the complicated table structure.

In addition, \textit{LLMs best attend to the table items when there are clues from both text and image.}
In \Cref{tab: highlight-vs-non-highlight}, we observe that marking the relevant cells on the image while mentioning them through text leads to the most correctly answered examples (61 examples by GPT-4 and 38 by Gemini$_\text{pro}$ at the last row in \Cref{tab: highlight-vs-non-highlight}).
Such a dual-modality approach that combines visual cues with text references, enhances LLMs' overall reasoning ability over the tabular data.

\subsection*{RQ5 and RQ6}
We include two additional research questions and our answer to them in \Cref{app-sec:rq-contd}, including whether these LLMs can reconstruct tables from the image, and whether multimodal LLMs benefit from having both formats simultaneously as input.

\begin{table}[t]
    \small
    \centering
    \begin{tabular}{cccrc}
    \toprule
         & \opensource & \closedsource &  \multicolumn{1}{c}{$\Delta$} & Metric\\
    \midrule
      FinQA  & 47.0 & 57.0 & +10.0 & \multirow{4}{*}{Acc}\\
      LogicNLG & 43.4 & 58.5  & +15.1 & \\
      TabFact & 51.8 & 74.7  & +22.9 & \\
      WikiTQ & 69.0 & 86.0 & +17.0 & \\
      \midrule
      E2E & 37.1 & 46.0 & + 8.9  & \multirow{2}{*}{ROUGE-L}\\
      ToTTo & 30.1 & 47.7 & +17.6 & \\
      \bottomrule
    \end{tabular}
    \caption{Performance scores of the best performed open-source (\opensource) LLM we test, Llama-2-70B versus closed-source (\closedsource) LLM we test, GPT-4 on different datasets.
    The closed-source LLMs always outperform the open-source LLMs and we report the performance difference $\Delta$ between them.
    For consistency across different datasets, we \textbf{do not include the performances with highlighting cells}\protect\footnotemark in this comparison.
    \Cref{tab:open-vs-close} in \Cref{app-sec:open-vs-close} provides what combinations of table representation and prompting method yield this performance.}
    \label{tab:open-vs-close-brief}
\end{table}

\footnotetext{Except for ToTTo, where the task is to generate the sentence based on the highlighted cells. On ToTTo, we include the highlight information just in text.}

\section{Open Problems to Increase the Performance of LLMs on Tabular Data}

\begin{figure}[t]
    \centering
    \includegraphics[width=0.95\linewidth]{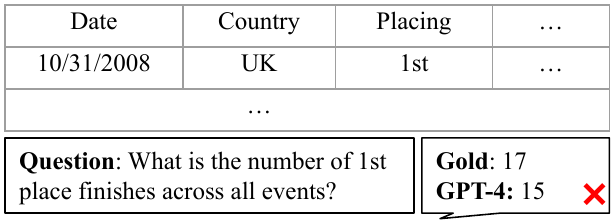}
    \caption{An example from WikiTQ where GPT-4 cannot answer it correctly with bracket table representation and chain-of-thought prompting.
    In addition, for most of the table representations and prompting styles, GPT-4 cannot answer this question correctly.}
    \label{fig:wikitq-gpt-4-cannot-answer}
\end{figure}

\begin{figure}[t]
    \centering
    \includegraphics[width=0.95\linewidth]{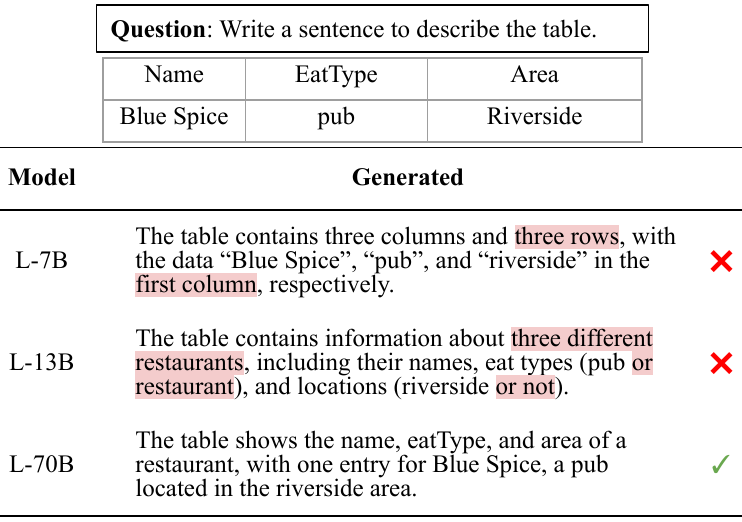}
    \caption{Table-to-text generation from E2E. 
    We use vanilla prompting and bracket table representation across all the models.
    ``L-7B/13B/70B'' represents Llama-2-7B/13B/70B, respectively.
    We highlight the part where the model generates incorrectly in red.}
    \label{fig:e2e_example_diff_llms}
\end{figure}

\paragraph{Mathematical reasoning.}
We observe that \textit{LLMs are not good at arithmetic reasoning} similar to the findings in prior works \cite{hendrycksmath2021, imani-etal-2023-mathprompter}.
As shown in \Cref{fig:wikitq-gpt-4-cannot-answer}, simple arithmetic computing like counting the total number of rows that satisfy certain conditions (`1st' in \Cref{fig:wikitq-gpt-4-cannot-answer}) still poses challenges even for GPT-4. 
\textit{This suggests that these previously proposed benchmarks are still valuable in evaluating LLMs}, as many of these datasets involve arithmetic reasoning such as FinQA and WikiTQ, which is challenging for even the most sophisticated LLMs like GPT-4.
For datasets like E2E for table-to-text generation, though relatively easy for advanced models like GPT-4, smaller models cannot describe the table fully correctly as shown in \Cref{fig:e2e_example_diff_llms}.
Therefore, these datasets are still valuable for benchmarking the progress of smaller LLMs.

\Cref{app-sec: dataset-examples} provides more examples from each dataset where LLMs can and cannot answer well.

\paragraph{Closing the gap between open-source and closed-source LLMs}

% \paragraph{Test:}
In \Cref{tab:open-vs-close-brief}, we report the best performance among different prompting methods for the best performed open-source Llama-2 models versus the best performed closed-source GPT models on FinQA, LogicNLG, TabFact, and WikiTQ.
We note that on these tasks, GPT-4 always performs the best among all the closed-source LLMs we test.
% We select the best performance among different prompting methods for comparison.

We observe that \textit{open-source Llama-2 models obtain decent performances across these benchmarks.} as shown in \Cref{tab:open-vs-close}.
% In \Cref{tab:open-vs-close}, open-source Llama-2 models obtain decent performances across these datasets.
\Cref{fig:e2e_example_diff_llms} shows an example from E2E dataset.
The smaller Llama-2 models such as Llama-2-7B or Llama-2-13B make mistakes in counting rows.
However, they capture almost all the information in the table including the name, eat type, and area of the restaurant.
As the model scales up, the Llama-2 70B model can describe the table accurately.

% \dnaihao{scaling up does help the model getting better}.

However, \textit{significant performance gaps exist between open-source Llama-2 models and closed-source GPT-4 models.}
% In \Cref{tab:open-vs-close}, we notice that GPT-4 obtains the best performance on all four datasets.
In \Cref{tab:open-vs-close-brief}, the gap between open-source Llama-2 models and GPT-4 can be as large as 15\% on FinQA and 22.9\% on TabFact. 
Even on LogicNLG which has the smallest performance gap, there is an 8.4\% difference between the Llama-2 and GPT models.
As Llama models often serve as the foundation models for a wide range of NLP research \cite{roziere2023code, xu2023lemur}, we need the effort from the open-source community to keep developing stronger LLMs to close the gap between open-source and closed-source LLMs.
% , suggesting the need for the open-source community to keep developing stronger LLMs.

\section{Conclusion}
% In conclusion, this paper has presented a comprehensive examination of the capabilities and limitations of large language models (LLMs) in processing and generating insights from structured data, such as tables. 
We have explored various representation strategies, including both text-based and innovative image-based approaches, to understand how to use LLMs effectively in tasks involving tabular data.
We demonstrate the effectiveness of image-based representations and reveal the impact of prompting strategies on the performance of LLMs. 
We believe our insights contribute to the understanding of LLMs and how to optimize LLMs for 
tabular data processing.

\section{Ethical Statement}
We conduct our studies on six pre-existing and publically available datasets using various existing LLMs.
Prior works have pointed out the potential bias in these LLMs \cite{bender2021dangers} which practitioners need to be aware of.

\section{Limitations}
In this study, we do not intend to exhaust every possible text representation, image representation of tables, or every possible LLM.
Moreover, we do not have access to the closed-source LLMs behind their API.
Instead, we designed our experiments to try to explore the research questions we raised in this paper.
We hope our findings and insights in this paper can inspire future research on table-related tasks.
% Entries for the entire Anthology, followed by custom entries

\section*{Acknowledgements}
We thank Inderjeet Nair for the early discussion on different forms of table representations; Alexander Hanbo Li, and Patrick Ng for discussions of our work and pointers to related work; the authors of the table-related datasets we use in our experiments for making their datasets publically available; the anonymous reviewers for their valuable suggestions.
This project was partially supported by the OpenAI Researcher Access Program and by a  Microsoft Foundational Model grant.

\bibliography{anthology,custom}

\appendix

\newpage

\section{Contributions}
\label{app-sec: contributions}

\paragraph{Idea Proposal.}
Naihao Deng proposed the idea of evaluating LLMs' performance on table-related tasks with various text-based and image-based table representations.

\paragraph{Background Survey.}
Zhenjie Sun conducted a careful survey on table-related tasks.
Naihao Deng did the initial background survey on table-related tasks when he was a research assistant in Westlake University with Yue Zhang.

\paragraph{Implementation.}
Zhenjie Sun came up with various table representations and designed all the prompts. Zhenjie Sun also implemented the evaluation pipeline for autonomous metrics.  Naihao Deng and Zhenjie Sun implemented the pipeline for human evaluation. 

\paragraph{Experiments.}
Zhenjie Sun and Naihao Deng conducted all the experiments.
Specifically, Zhenjie Sun conducted experiments on GPT-3.5, GPT-4 with text-based table representations,  and Gemini$_\text{pro}$.
Naihao Deng conducted experiments on Llama-2 models and GPT-4 with image-based table representations.

\paragraph{Result Aggregation.}
Zhenjie Sun conducted the result aggregation for all the experiments.

\paragraph{Human Evaluation.}
Zhenjie Sun, Naihao Deng, Ruiqi He, Aman Sikka conducted the human evaluation for the model generation.

\paragraph{Paper Writing.}
Naihao Deng drafted the paper.
Zhenjie Sun drafted the prompting strategies and metrics, and actively got involved in discussions of result analysis.
Naihao Deng and Zhenjie Sun plotted all the tables and figures.
Naihao Deng, Zhenjie Sun selected examples that appeared in this paper.
Ruiqi He, Aman Sikka provided suggestions on example selections.
Rada Mihalcea, Yue Zhang, Lin Ma, and Yulong Chen participated in discussions throughout the entire project and provided revisions and feedback on the paper.

\section{Statistical Significance Test Results}
\label{app-sec:significance-test}

We have observed significant differences between LLMs’ performance.
We conduct three statistical significance tests, including Fisher's Exact test, McNemar’s Test, and proportion Z test for the model predictions. 

With p < 0.05:
\begin{enumerate}
    \item For Fisher’s Exact test, we find a statistically significant difference in GPT-4 performance between T and V inputs for FinQA, LogicNLG, TabFact, WikiTQ (\Cref{fig:text_vision_comparison}), its performance between vanilla and expert prompting for table text input (\Cref{tab:rq2-vanilla-bracket}), and its performance difference under vanilla and chain-of-thought prompting for image-based table representations (\Cref{tab:image-prompting-rq3}).
    \item For McNemar’s Test, we find statistically significant differences between GPT-4 performance between T and V inputs for FinQA, TabFact, and WikiTQ (\Cref{fig:text_vision_comparison}), as well as between vanilla and expert prompting for table text input (\Cref{tab:rq2-vanilla-bracket}).
    \item For the proportion Z test, we find a statistically significant difference in GPT-4 performance between T and V inputs for FinQA, WikiTQ (\Cref{fig:text_vision_comparison}).
\end{enumerate}

\section{Research Questions Cont'd}
\label{app-sec:rq-contd}

\subsection{RQ1 Cont'd. Can we use image-based representations of tabular data?}
\label{app-subsec:rq1-contd}
\begin{figure*}[t]
\centering
\includegraphics[width=0.95\linewidth]{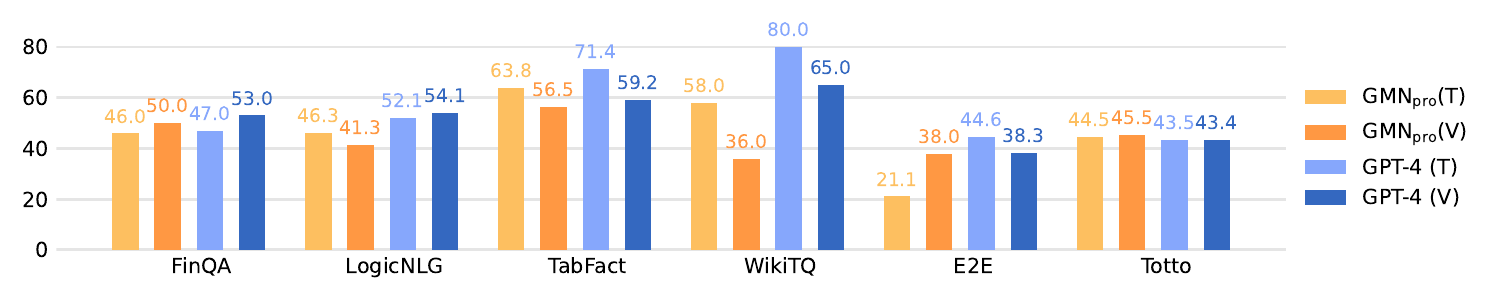} 
\caption{Performance comparison between passing the text versus image representations of tables to GPT-4 and Gemini$_\text{Pro}$ across FinQA, LogicNLG, TabFact, and WikiTQ by accuracy, and E2E and ToTTo by ROUGE-L scores.
We use the linearized table (Vanilla-T) as the text-based representation, the original table image (Vanilla-V) as the image-based representation, and CoT prompting.}
\label{fig:text_vision_comparison_cot}
\end{figure*}
\begin{figure*}[t]
\centering
\includegraphics[width=0.95\linewidth]{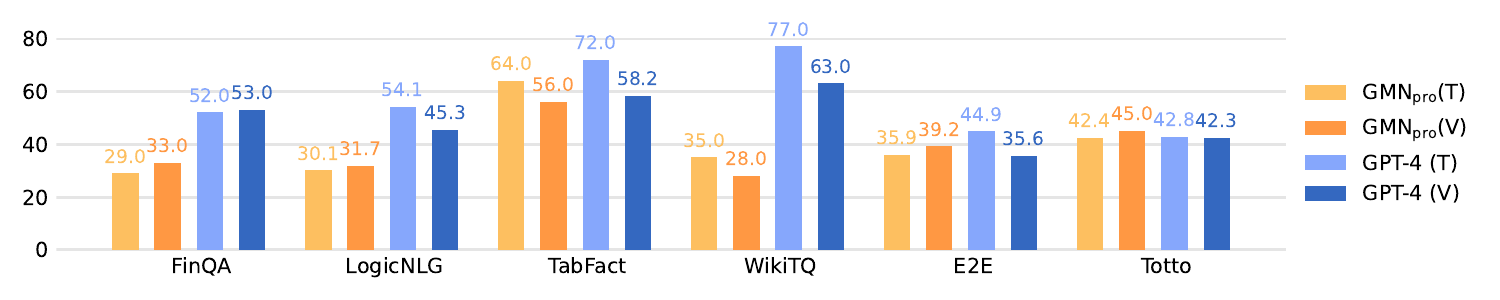} 
\caption{Performance comparison between passing the text versus image representations of tables to GPT-4 and Gemini$_\text{Pro}$ across FinQA, LogicNLG, TabFact, and WikiTQ by accuracy, and E2E and ToTTo by ROUGE-L scores.
We use the linearized table (Vanilla-T) as the text-based representation, the original table image (Vanilla-V) as the image-based representation, and expert prompting.}
\label{fig:text_vision_comparison_expert}
\end{figure*}

\Cref{fig:text_vision_comparison_cot} and \Cref{fig:text_vision_comparison_expert} show the performance comparison between feeding text representations versus image representations of the table to GPT-4 and Gemini$_\text{pro}$ for chain-of-thought and expert prompting, respectively.
The results resemble similar trends as \Cref{fig:text_vision_comparison}.

\subsection{RQ2 Cont'd. How do different text-based prompt methods affect LLMs' performance on tabular-related tasks?}
\label{app-subsec:rq2-contd}

\begin{table}[t]
    \small
    \centering
    \setlength{\tabcolsep}{3pt}
    \begin{tabular}{ccccccc}
               \toprule
        & \multicolumn{2}{c}{GPT} & \multirow{2}{*}{GMN$_\text{pro}$} & \multicolumn{3}{c}{Llama-2} \\
        &3.5 & 4& & 7B & 13B & 70B\\
\midrule
\multicolumn{2}{l}{\textbf{Vanilla-T}} & & & & & \\
V & \applyGradientc{52.5} & \applyGradientc{60.3} & \applyGradientc{37.1} & \applyGradientc{28.8} & \applyGradientc{35.3} & \applyGradientc{42.7}\\
 E & \applyGradientc{51.0} & \applyGradientc{63.8} & \applyGradientc{39.5} & \applyGradientc{29.0} & \applyGradientc{35.1} & \applyGradientc{46.7} \\
CoT & \applyGradientc{55.2} & \applyGradientc{62.6} & \applyGradientc{53.5} & \applyGradientc{32.1} & \applyGradientc{37.6} & \applyGradientc{48.3} \\
\midrule
\multicolumn{2}{l}{\textbf{Bracket}} & & & & & \\
V & \applyGradientc{50.9} & \applyGradientc{60.1} & \applyGradientc{38.4} & \applyGradientc{28.4} & \applyGradientc{36.6} & \applyGradientc{42.2} \\
 E & \applyGradientc{47.9} & \applyGradientc{62.8} & \applyGradientc{39.5} & \applyGradientc{28.1} & \applyGradientc{34.5} & \applyGradientc{45.8} \\
CoT & \applyGradientc{51.4} & \applyGradientc{61.9} & \applyGradientc{57.3} & \applyGradientc{34.2} & \applyGradientc{39.3} & \applyGradientc{50.0} \\
\midrule
\multicolumn{2}{l}{\textbf{Column-JSON}} & & & & & \\
V & \applyGradientc{48.3} & \applyGradientc{59.5} & \applyGradientc{32.6} & \applyGradientc{24.9} & \applyGradientc{28.8} & \applyGradientc{39.2}  \\
 E & \applyGradientc{48.8} & \applyGradientc{62.8} & \applyGradientc{34.0} & \applyGradientc{26.4} & \applyGradientc{28.2} & \applyGradientc{42.5} \\
CoT & \applyGradientc{51.2} & \applyGradientc{59.6} & \applyGradientc{53.6} & \applyGradientc{28.0} & \applyGradientc{34.8} & \applyGradientc{42.8} \\
\midrule
\multicolumn{2}{l}{\textbf{Row-JSON}} & & & & & \\
V & \applyGradientc{49.7} & \applyGradientc{62.3} & \applyGradientc{41.2} & \applyGradientc{27.9} & \applyGradientc{32.6} & \applyGradientc{40.9} \\
 E & \applyGradientc{53.7} & \applyGradientc{63.8} & \applyGradientc{39.4} & \applyGradientc{26.4} & \applyGradientc{31.6} & \applyGradientc{45.4}  \\
CoT & \applyGradientc{53.3} & \applyGradientc{62.0} & \applyGradientc{52.1} & \applyGradientc{31.0} & \applyGradientc{35.7} & \applyGradientc{48.4} \\
\midrule
\multicolumn{2}{l}{\textbf{Row-Identifier}} & & & & &\\
V & \applyGradientc{52.0} & \applyGradientc{61.2} & \applyGradientc{38.6} & \applyGradientc{27.9} & \applyGradientc{38.5} & \applyGradientc{43.2} \\
 E & \applyGradientc{53.2} & \applyGradientc{63.0} & \applyGradientc{38.2} & \applyGradientc{26.1} & \applyGradientc{34.0} & \applyGradientc{41.8} \\
CoT & \applyGradientc{51.6} & \applyGradientc{62.1} & \applyGradientc{56.5} & \applyGradientc{30.6} & \applyGradientc{33.0} & \applyGradientc{45.9} \\

        \bottomrule
    \end{tabular}
    \caption{For text-based table representations, averaged accuracy scores across FinQA, LogicNLG, TabFact, and WikiTQ for different LLMs.
    ``GMN$_\text{pro}$'' represents Gemini$_\text{pro}$ model, ``V'', ``E'', and ``CoT'' represent vanilla, expert and chain-of-thought prompting, respectively.}
    \label{tab:rq2-all}
\end{table}

\begin{table}[t]
    \small
    \centering
    \setlength{\tabcolsep}{3pt}
    \begin{tabular}{ccccccccc}
               \toprule
        & \multicolumn{2}{c}{GPT} & \multirow{2}{*}{GMN$_\text{pro}$} & \multicolumn{3}{c}{Llama-2}  \\
        &3.5 & 4& & 7B & 13B & 70B \\
% \midrule
% \multicolumn{2}{l}{\textbf{Vanilla-T}} & & & & & \\
% V & 28.6 & 44.8 & 35.6 & 21.2 & 20.6 & 20.0  \\
% E & 29.3 & 44.9 & 35.9 & 15.8 & 20.6 & 16.8 \\
% CoT & 21.3 & 44.6 & 21.1 & 16.7 & 17.7 & 18.1\\
% \midrule
% \multicolumn{2}{l}{\textbf{Bracket}} & & & & & \\
% V & 42.0 & 45.2 & 26.1 & 23.0 & 23.7 & 21.2 \\
% E & 41.7 & 43.2 & 29.6 & 19.4 & 24.7 & 21.4  \\
% CoT & 31.3 & 42.4 & 19.5 & 18.8 & 21.0 & 18.3  \\
% \midrule
% \multicolumn{2}{l}{\textbf{Column-JSON}} & & & & & \\
% V & 45.5 & 45.6 & 41.6 & 37.1 & 26.2 & 31.4 \\
% E & 43.5 & 45.1 & 41.9 & 27.2 & 25.4 & 29.0\\
% CoT & 43.7 & 46.0 & 22.3 & 30.0 & 23.4 & 23.2 \\
% \midrule
% \multicolumn{2}{l}{\textbf{Row-JSON}} & & & & &\\
% V & 44.6 & 45.7 & 28.8 & 32.4 & 21.5 & 25.9  \\
% E & 43.3 & 45.0 & 21.9 & 27.6 & 28.0 & 27.5 \\
% CoT & 43.6 & 44.7 & 22.1 & 27.0 & 24.1 & 19.3\\
        \midrule
        \multicolumn{2}{l}{\textbf{Vanilla-T}} & & & & & \\
        V & \applyGradientd{28.6} & \applyGradientd{44.8} & \applyGradientd{35.6} & \applyGradientd{21.2} & \applyGradientd{20.6} & \applyGradientd{20.0} \\
        E & \applyGradientd{29.3} & \applyGradientd{44.9} & \applyGradientd{35.9} & \applyGradientd{15.8} & \applyGradientd{20.6} & \applyGradientd{16.8} \\
        CoT & \applyGradientd{21.3} & \applyGradientd{44.6} & \applyGradientd{21.1} & \applyGradientd{16.7} & \applyGradientd{17.7} & \applyGradientd{18.1} \\
        \midrule
        \multicolumn{2}{l}{\textbf{Bracket}} & & & & & \\
        V & \applyGradientd{42.0} & \applyGradientd{45.2} & \applyGradientd{26.1} & \applyGradientd{23.0} & \applyGradientd{23.7} & \applyGradientd{21.2} \\
        E & \applyGradientd{41.7} & \applyGradientd{43.2} & \applyGradientd{29.6} & \applyGradientd{19.4} & \applyGradientd{24.7} & \applyGradientd{21.4} \\
        CoT & \applyGradientd{31.3} & \applyGradientd{42.4} & \applyGradientd{19.5} & \applyGradientd{18.8} & \applyGradientd{21.0} & \applyGradientd{18.3} \\
        \midrule
        \multicolumn{2}{l}{\textbf{Column-JSON}} & & & & & \\
        V & \applyGradientd{45.5} & \applyGradientd{45.6} & \applyGradientd{41.6} & \applyGradientd{37.1} & \applyGradientd{26.2} & \applyGradientd{31.4} \\
        E & \applyGradientd{43.5} & \applyGradientd{45.1} & \applyGradientd{41.9} & \applyGradientd{27.2} & \applyGradientd{25.4} & \applyGradientd{29.0} \\
        CoT & \applyGradientd{43.7} & \applyGradientd{46.0} & \applyGradientd{22.3} & \applyGradientd{30.0} & \applyGradientd{23.4} & \applyGradientd{23.2} \\
        \midrule
        \multicolumn{2}{l}{\textbf{Row-JSON}} & & & & & \\
        V & \applyGradientd{44.6} & \applyGradientd{45.7} & \applyGradientd{28.8} & \applyGradientd{32.4} & \applyGradientd{21.5} & \applyGradientd{25.9} \\
        E & \applyGradientd{43.3} & \applyGradientd{45.0} & \applyGradientd{21.9} & \applyGradientd{27.6} & \applyGradientd{28.0} & \applyGradientd{27.5} \\
        CoT & \applyGradientd{43.6} & \applyGradientd{44.7} & \applyGradientd{22.1} & \applyGradientd{27.0} & \applyGradientd{24.1} & \applyGradientd{19.3} \\
                \bottomrule
    \end{tabular}
    \caption{For text-based table representations, ROUGE-L scores on E2E for different LLMs.
    ``GMN$_\text{pro}$'' represents Gemini$_\text{pro}$ model, ``V'', ``E'', and ``CoT'' represent vanilla, expert and chain-of-thought prompting, respectively.
    We do not include the Row-Identifier here as all the tables in E2E dataset only contains one row other than the header row.}
    \label{tab:rq2-e2e-rougeL}
\end{table}

\begin{table}[t]
    \small
    \centering
    \setlength{\tabcolsep}{2.5pt}
    \begin{tabular}{ccccccc}
        \toprule
        & \multicolumn{2}{c}{GPT} & \multirow{2}{*}{GMN$_\text{pro}$} & \multicolumn{3}{c}{Llama-2} \\
        &3.5 & 4& & 7B & 13B & 70B \\

        \midrule
        \multicolumn{2}{l}{\textbf{Vanilla-T}} & & & & & \\
        V & \applyGradientf{28} & \applyGradientf{79} & \applyGradientf{60} & \applyGradientf{23} & \applyGradientf{15} & \applyGradientf{24} \\
        E & \applyGradientf{26} & \applyGradientf{81} & \applyGradientf{50} & \applyGradientf{8} & \applyGradientf{22} & \applyGradientf{12} \\
        CoT & \applyGradientf{23} & \applyGradientf{86} & \applyGradientf{32} & \applyGradientf{17} & \applyGradientf{14} & \applyGradientf{19} \\
        \midrule
        \multicolumn{2}{l}{\textbf{Bracket}} & & & & & \\
        V & \applyGradientf{90} & \applyGradientf{94} & \applyGradientf{33} & \applyGradientf{28} & \applyGradientf{69} & \applyGradientf{32} \\
        E & \applyGradientf{94} & \applyGradientf{92} & \applyGradientf{39} & \applyGradientf{29} & \applyGradientf{78} & \applyGradientf{34} \\
        CoT & \applyGradientf{74} & \applyGradientf{94} & \applyGradientf{36} & \applyGradientf{26} & \applyGradientf{62} & \applyGradientf{37} \\
        \midrule
        \multicolumn{2}{l}{\textbf{Column-JSON}} & & & & & \\
        V & \applyGradientf{91} & \applyGradientf{82} & \applyGradientf{88} & \applyGradientf{63} & \applyGradientf{63} & \applyGradientf{77} \\
        E & \applyGradientf{91} & \applyGradientf{84} & \applyGradientf{85} & \applyGradientf{56} & \applyGradientf{76} & \applyGradientf{73} \\
        CoT & \applyGradientf{90} & \applyGradientf{84} & \applyGradientf{60} & \applyGradientf{55} & \applyGradientf{67} & \applyGradientf{75} \\
        \midrule
        \multicolumn{2}{l}{\textbf{Row-JSON}} & & & & & \\
        V & \applyGradientf{94} & \applyGradientf{93} & \applyGradientf{57} & \applyGradientf{54} & \applyGradientf{41} & \applyGradientf{48} \\
        E & \applyGradientf{94} & \applyGradientf{94} & \applyGradientf{33} & \applyGradientf{58} & \applyGradientf{72} & \applyGradientf{65} \\
        CoT & \applyGradientf{95} & \applyGradientf{96} & \applyGradientf{62} & \applyGradientf{62} & \applyGradientf{49} & \applyGradientf{60} \\
        
                \bottomrule
    \end{tabular}
    \caption{For text-based table representations, manual annotation scores (whether the generation contains all the information from the table without any additional or mis-information) on E2E for all LLMs.
    ``GMN$_\text{pro}$'' represents Gemini$_\text{pro}$ model, ``V'', ``E'', and ``CoT'' represent vanilla, expert and chain-of-thought prompting, respectively.}
    \label{tab:rq2-e2e-table-info}
\end{table}

\begin{table}[t]
    \small
    \centering
    \setlength{\tabcolsep}{3pt}
    \begin{tabular}{ccccccc}
               \toprule
        & \multicolumn{2}{c}{GPT} & \multirow{2}{*}{GMN$_\text{pro}$} & \multicolumn{3}{c}{Llama-2} \\
        &3.5 & 4& & 7B & 13B & 70B\\

        % \midrule
        % \multicolumn{2}{l}{\textbf{Vanilla-T}} & & & & &\\
        % V & 43.3 & 43.9 & 42.4 & 21.6 & 22.9 & 27.2 \\
        % E & 41.4 & 42.8 & 42.4 & 20.7 & 22.2 & 26.3 \\
        % CoT & 42.7 & 43.5 & 44.5 & 19.8 & 22.8 & 27.4 \\
        % \midrule
        % \multicolumn{2}{l}{\textbf{Bracket}} & & & & &\\
        % V & 44.2 & 44.9 & 44.8 & 23.6 & 24.8 & 28.9 \\
        % E & 41.7 & 43.0 & 44.1 & 22.3 & 23.1 & 29.1 \\
        % CoT & 43.9 & 45.5 & 43.9 & 23.3 & 24.0 & 29.6 \\
        % \midrule
        % \multicolumn{2}{l}{\textbf{Column-JSON}} & & & & &\\
        % V & 45.5 & 45.5 & 44.5 & 22.1 & 22.7 & 30.1 \\
        % E & 41.1 & 43.1 & 44.8 & 19.9 & 20.8 & 10.3 \\
        % CoT & 43.9 & 44.2 & 45.1 & 22.1 & 21.5 & 28.8 \\
        % \midrule
        % \multicolumn{2}{l}{\textbf{Row-JSON}} & & & & & \\
        % V & 43.1 & 45.1 & 43.7 & 22.3 & 21.9 & 29.4 \\
        % E & 40.8 & 43.1 & 43.4 & 21.5 & 21.4 & 27.0 \\
        % CoT & 42.3 & 45.2 & 45.1 & 22.9 & 22.5 & 26.9 \\
        % \midrule
        % \multicolumn{2}{l}{\textbf{Row-Identifier}} & & & & &\\
        % V & 42.2 & 44.6 & 43.2 & 19.1 & 22.2 & 28.3 \\
        % E & 40.1 & 42.7 & 42.5 & 21.3 & 21.3 & 26.5 \\
        % CoT & 42.0 & 43.7 & 44.2 & 19.1 & 22.3 & 28.0 \\
        \midrule
        \multicolumn{2}{l}{\textbf{Vanilla-T}} & & & & & \\
        V & \applyGradientg{43.3} & \applyGradientg{43.9} & \applyGradientg{42.4} & \applyGradientg{21.6} & \applyGradientg{22.9} & \applyGradientg{27.2} \\
        E & \applyGradientg{41.4} & \applyGradientg{42.8} & \applyGradientg{42.4} & \applyGradientg{20.7} & \applyGradientg{22.2} & \applyGradientg{26.3} \\
        CoT & \applyGradientg{42.7} & \applyGradientg{43.5} & \applyGradientg{44.5} & \applyGradientg{19.8} & \applyGradientg{22.8} & \applyGradientg{27.4} \\
        \midrule
        \multicolumn{2}{l}{\textbf{Bracket}} & & & & & \\
        V & \applyGradientg{44.2} & \applyGradientg{44.9} & \applyGradientg{44.8} & \applyGradientg{23.6} & \applyGradientg{24.8} & \applyGradientg{28.9} \\
        E & \applyGradientg{41.7} & \applyGradientg{43.0} & \applyGradientg{44.1} & \applyGradientg{22.3} & \applyGradientg{23.1} & \applyGradientg{29.1} \\
        CoT & \applyGradientg{43.9} & \applyGradientg{45.5} & \applyGradientg{43.9} & \applyGradientg{23.3} & \applyGradientg{24.0} & \applyGradientg{29.6} \\
        \midrule
        \multicolumn{2}{l}{\textbf{Column-JSON}} & & & & & \\
        V & \applyGradientg{45.5} & \applyGradientg{45.5} & \applyGradientg{44.5} & \applyGradientg{22.1} & \applyGradientg{22.7} & \applyGradientg{30.1} \\
        E & \applyGradientg{41.1} & \applyGradientg{43.1} & \applyGradientg{44.8} & \applyGradientg{19.9} & \applyGradientg{20.8} & \applyGradientg{10.3} \\
        CoT & \applyGradientg{43.9} & \applyGradientg{44.2} & \applyGradientg{45.1} & \applyGradientg{22.1} & \applyGradientg{21.5} & \applyGradientg{28.8} \\
        \midrule
        \multicolumn{2}{l}{\textbf{Row-JSON}} & & & & & \\
        V & \applyGradientg{43.1} & \applyGradientg{45.1} & \applyGradientg{43.7} & \applyGradientg{22.3} & \applyGradientg{21.9} & \applyGradientg{29.4} \\
        E & \applyGradientg{40.8} & \applyGradientg{43.1} & \applyGradientg{43.4} & \applyGradientg{21.5} & \applyGradientg{21.4} & \applyGradientg{27.0} \\
        CoT & \applyGradientg{42.3} & \applyGradientg{45.2} & \applyGradientg{45.1} & \applyGradientg{22.9} & \applyGradientg{22.5} & \applyGradientg{26.9} \\
        \midrule
        \multicolumn{2}{l}{\textbf{Row-Identifier}} & & & & & \\
        V & \applyGradientg{42.2} & \applyGradientg{44.6} & \applyGradientg{43.2} & \applyGradientg{19.1} & \applyGradientg{22.2} & \applyGradientg{28.3} \\
        E & \applyGradientg{40.1} & \applyGradientg{42.7} & \applyGradientg{42.5} & \applyGradientg{21.3} & \applyGradientg{21.3} & \applyGradientg{26.5} \\
        CoT & \applyGradientg{42.0} & \applyGradientg{43.7} & \applyGradientg{44.2} & \applyGradientg{19.1} & \applyGradientg{22.3} & \applyGradientg{28.0} \\
                \bottomrule
    \end{tabular}
    \caption{For text-based table representations, ROUGE-L scores on ToTTo for all LLMs.
    ``GMN$_\text{pro}$'' represents Gemini$_\text{pro}$ model, ``V'', ``E'', and ``CoT'' represent vanilla, expert and chain-of-thought prompting, respectively.}
    \label{tab:rq2-totto-rougeL}
\end{table}

\Cref{tab:rq2-all} reports the averaged accuracy scores across FinQA, LogicNLG, TabFact and WikiTQ that use accuracy as the metric.
\Cref{tab:rq2-e2e-rougeL} and \Cref{tab:rq2-totto-rougeL} report the ROUGE-L scores of LLMs' generation on E2E and ToTTo dataset, respectively.
\Cref{tab:rq2-e2e-table-info} reports the scores annotated manually by the authors.
As discussed in \Cref{subsec: metrics}, the authors manually check whether the generated sentence captures all the information from the table and does not include any additional or misinformation.
We assign ``1'' for sentences who satisfy the criteria and ``0'' otherwise.

\subsection{RQ3 Cont'd. How do different image-based prompt methods affect LLMs' performance on tabular-related tasks?}
\label{app-subsec:rq3-contd}

\begin{table}[t]
    \small
    \centering
    \begin{tabular}{ccccccc}
    \toprule
         &  \multicolumn{3}{c}{GPT-4} & \multicolumn{3}{c}{Gemini$_\text{pro}$}\\
         & V  & E & CoT & V & E & CoT \\
    \midrule
    
      VV & \applyGradienti{44.0} & \applyGradienti{42.3} & \applyGradienti{43.4} & \applyGradienti{46.0} & \applyGradienti{45.0} & \applyGradienti{45.5} \\
CC & \applyGradienti{44.8} & \applyGradienti{41.7} & \applyGradienti{44.1} & \applyGradienti{47.7} & \applyGradienti{44.8} & \applyGradienti{45.1} \\
RC & \applyGradienti{44.5} & \applyGradienti{42.8} & \applyGradienti{43.7} & \applyGradienti{46.3} & \applyGradienti{44.6} & \applyGradienti{45.0} \\
    \bottomrule 
    \end{tabular}
    \caption{For image-based table representations, ROUGE-L scores on E2E for GPT-4 and Gemini$_\text{pro}$.
    For the headers, ``V'', ``E'', and ''CoT'' represent vanilla, expert, and chain-of-thought prompting, respectively.
    For the row names, ``VV'', ``CC', and ``RC'' represent Vanilla-V, Column-Color, and Row-Color, respectively.}
    \label{tab:image-prompting-e2e-rougeL}
\end{table}

\begin{table}[t]
    \small
    \centering
    \begin{tabular}{ccccccc}
    \toprule
         &  \multicolumn{3}{c}{GPT-4} & \multicolumn{3}{c}{Gemini$_\text{pro}$}\\
         & V  & E & CoT & V & E & CoT \\
    \midrule
    
VV & \applyGradientj{44.0} & \applyGradientj{42.3} & \applyGradientj{43.4} & \applyGradientj{46.0} & \applyGradientj{45.0} & \applyGradientj{45.5} \\
CC & \applyGradientj{44.8} & \applyGradientj{41.7} & \applyGradientj{44.1} & \applyGradientj{47.7} & \applyGradientj{44.8} & \applyGradientj{45.1} \\
RC & \applyGradientj{44.5} & \applyGradientj{42.8} & \applyGradientj{43.7} & \applyGradientj{46.3} & \applyGradientj{44.6} & \applyGradientj{45.0} \\

    \bottomrule 
    \end{tabular}
    \caption{For image-based table representations, ROUGE-L scores on ToTTo for GPT-4 and Gemini$_\text{pro}$.
    For the headers, ``V'', ``E'', and ''CoT'' represent vanilla, expert, and chain-of-thought prompting, respectively.
    For the row names, ``VV'', ``CC', and ``RC'' represent Vanilla-V, Column-Color, and Row-Color, respectively.}
    \label{tab:image-prompting-totto-rougeL}
\end{table}

\begin{table}[t]
    \small
    \centering
    \begin{tabular}{ccccccc}
    \toprule
         &  \multicolumn{3}{c}{GPT-4} & \multicolumn{3}{c}{Gemini$_\text{pro}$}\\
         & V  & E & CoT & V & E & CoT \\
    \midrule
        
VV & \applyGradientf{86} & \applyGradientf{83} & \applyGradientf{90} & \applyGradientf{77} & \applyGradientf{74} & \applyGradientf{78} \\
CC & \applyGradientf{86} & \applyGradientf{69} & \applyGradientf{93} & \applyGradientf{70} & \applyGradientf{61} & \applyGradientf{72} \\
RC & \applyGradientf{85} & \applyGradientf{70} & \applyGradientf{89} & \applyGradientf{61} & \applyGradientf{57} & \applyGradientf{60} \\

    \bottomrule 
    \end{tabular}
    \caption{For image-based table representations, manual annotation scores (whether the generation contains all the information from the table without any additional or mis-information) on E2E for GPT-4 and Gemini$_\text{pro}$.
    For the headers, ``V'', ``E'', and ''CoT'' represent vanilla, expert, and chain-of-thought prompting, respectively.
    For the row names, ``VV'', ``CC', and ``RC'' represent Vanilla-V, Column-Color, and Row-Color, respectively.}
    \label{tab:image-prompting-e2e-table-info}
\end{table}
\begin{table}[t]
    \small
    \centering
    \begin{tabular}{cccc}
    \toprule
         &  GPT-4 &Gemini$_\text{pro}$ & Metric\\
        
    \midrule
    WikiTQ & 80.0 & 58.0 & \multirow{4}{*}{Acc} \\
    TabFact & 64.0 & 61.3\\
     LogicNLG & 48.0 & 33.3 \\
     FinQA  & 61.0 & 45.0 \\
     \midrule
     ToTTo & 42.4 & 44.2 &\multirow{2}{*}{ROUGE-L}\\
       E2E & 42.6 & 41.1 \\ 
    
    \bottomrule 
    \end{tabular}
    \caption{Results on experiments providing both vanilla image and text representations of tables for GPT-4 and Gemini$_\text{pro}$.
   }
    \label{tab:Both-text-and-image}
\end{table}

\Cref{tab:image-prompting-e2e-rougeL,tab:image-prompting-totto-rougeL} report the ROUGE-L scores of GPT-4 and Gemini$_\text{pro}$ when we use image representations of tables on E2E and ToTTo dataset, respectively.
\Cref{tab:image-prompting-e2e-table-info} reports the scores annotated manually by the authors.

\subsection{RQ5: Can These LLMs reconstruct tables from the image?}
\paragraph{Test:} We conduct experiments to explore table reconstruction ideas using GPT4 on E2E and FinQA datasets.
\paragraph{TL;DR Answer:} Mostly yes.
\paragraph{Full Answer:} GPT4 successfully reconstructs 97/100 tables on the E2E dataset, and 72/100 tables on the FinQA dataset.
We notice that for simpler tables, GPT4 can reconstruct tables almost perfectly. For instance, in E2E, the only ``mistake'' the model makes is to output ``é'' instead of ``e'' in ``Rainbow Vegetarian Café''.

GPT4 also demonstrates a decent capability of reconstructing complicated tables. 
On FinQA, GPT-4 manages to reconstruct a 4-row$\times$10-column table with over 100 words (\Cref{tab:finqa-gpt4-recon}).

\begin{table*}
     \small
    \centering
    \begin{tabular}{cccc}
    \toprule
    Year Ended December 31 (in millions except rates) & 2018 & 2017 & 2016 \\
    \midrule
    Less: Average CIB Markets Interest-Earning Assets (c) & 609,635 & 540,835 & 520,307\\
    Average Interest-Earning Assets Excluding CIB Markets & \$1,619,553 & \$1,639,757 &	\$1,581,297 \\
    Net Interest Yield on Average Interest-Earning ... & 2.50\% (2.50\%)	&2.36\% (2.36\%) &	2.25\% (2.25\%) \\
    Net Interest Yield on Average CIB Markets ...	& 0.51	&0.86	& 1.22\\
    Net Interest Yield on Average Interest-Earning ...	& 3.25\% (3.25\%)	& 2.85\% (2.85\%)	& 2.59\% (2.59\%)\\
    ... & ...\\
    \bottomrule
    \end{tabular}
    \caption{On FinQA, GPT-4 manages to reconstruct this 4-row$\times$10-column table with over 100 words from its screenshot (we omit some rows to save space).}
    \label{tab:finqa-gpt4-recon}
\end{table*}

But GPT-4 is more prone to hallucination or messing up with the spatial relations when the table gets more complicated. 
For instance, for \Cref{tab:finqa-fail-recon}, GPT-4 reconstructs it as \Cref{tab:finqa-fail-recon-result}.
As expected, GPT-4 fails to answer the corresponding question to this table when using a table image as the input. 
However, we notice that when using the text representation of the table, GPT-4 also fails to answer this question.
This aligns with what we have reported in.

\begin{table*}
     \small
    \centering
     \begin{tabular}{cccc}
    \toprule
    Year Ended December 31 (in millions)&	2010	&2009	&2008\\
    \midrule
    U.S.	&\$16,568	&\$6,263	&\$-2,094 (2094)\\
    Non-U.S. (a)	&8,291	&9,804	&4,867\\
    Income Before Income Tax Expense/(Benefit) and Extraordinary Gain	&\$24,859	&\$16,067&	\$2,773\\
    \bottomrule
    \end{tabular}
    \caption{The table from FinQA where GPT-4 fails to reconstruct.}
    \label{tab:finqa-fail-recon}
\end{table*}

\begin{table*}
     \small
    \centering
     \begin{tabular}{cccccc}
    \toprule
    Year Ended December 31 (in millions) &	U.S.	&Non-U.S. (a)	&2010	&2009	&2008\\
    \midrule
   Income before income tax expense/(benefit)...	&\$16,568	&\$6,263	&\$-2,094 (2094)	&\$8,291	&\$9,804\\
&&&\$24,859	&\$16,067	&\$2,773\\
    \bottomrule
    \end{tabular}
    \caption{The reconstructed table from GPT4 (the correct table is in \Cref{tab:finqa-fail-recon} 
    We omit the content in one cell to save space.
    We notice that GPT4 messes up with the spatial relations in the table.}
    \label{tab:finqa-fail-recon-result}
\end{table*}

As expected, GPT-4 fails to answer the corresponding question to this table when using a table image as the input. 
However, we notice that when using the text representation of the table, GPT-4 also fails to answer this question.
This aligns with what we have reported in \Cref{fig:text_vision_comparison} that On FinQA, GPT-4 better leverages the image representations than the text representations in general. 

\subsection{RQ6: Do multimodal LLMs benefit from having both formats simultaneously as input?}
\paragraph{Test:} We test GPT-4 with vanilla prompting for the six datasets.

\paragraph{TL;DR Answer:} Not generally true.

\paragraph{Full Answer:}
As shown in \Cref{tab:both-repr}, it is not generally true that multimodal LLMs benefit from having both formats simultaneously as input.

There are cases where passing both image and text representations would increase the performance (e.g. GPT-4 on WikiTQ and FinQA). 
In other cases, the performance is comparable to either passing tables as text or image representation or lies in between.

\begin{table}
\small
\begin{tabular}{crrrc}
\toprule
     Datasets	&T+V	&T	&V  & Metric\\
     \midrule
     WikiTQ &80.0	&75.0	&60.0 & \multirow{4}{*}{Acc}\\
     TabFact	&64.0	&71.6	&60.5\\
     LogicNLG	&48.0	&51.5	&54.1\\
     FinQA	   &61.0	&43.0	&57.0 \\
     \midrule
     ToTTo &42.4	&43.9	&44.0 &\multirow{2}{*}{ROUGE-L}\\
    E2E    &42.6	&44.8	&39.4\\
\bottomrule
\end{tabular}
\caption{GPT-4's performance when we pass the text representation (T), image representation (V) and both representation (T+V) to the model.}
\label{tab:both-repr}
\end{table}

\section{Comparison of Llama Models and GPT-4 Models}
\label{app-sec:open-vs-close}

\begin{table*}[t]
    \small
    \centering
    \begin{tabular}{cccccrcccc}
    \toprule
         & Table Repr & Prompting & \opensource & \closedsource &  \multicolumn{1}{c}{$\Delta$} & Table Repr & Prompting & Metric \\
    \midrule
      FinQA  & Vanilla-T & CoT & 47.0 & 57.0 & +10.0 & Vanilla-V & Vanilla & \multirow{4}{*}{Acc}\\
      LogicNLG & Vanilla-T & CoT & 43.4 & 58.5  & +15.1 & Row-Color & CoT \\
      TabFact & Column-JSON & Expert & 51.8 & 74.7  & +22.9 & Row-JSON & Expert \\
      WikiTQ & Row-JSON & CoT & 69.0 & 86.0 & +17.0 & Row-Identifier & CoT \\
        \midrule
        E2E & Column-JSON & Vanilla & 37.1  & 46.0 & +8.9 & Column-JSON & CoT & \multirow{2}{*}{ROUGE-L}\\
        ToTTo & Column-JSON & Vanilla & 30.1  & 47.7 & +17.6 & Column-Color & Vanilla \\
      \bottomrule
    \end{tabular}
    \caption{Performance scores of the best performed open-source (\opensource) LLM we test, Llama-2-70B versus closed-source (\closedsource) LLM we test, GPT-4 on different datasets. Four datasets uses accuracy as metrics and two datasets (E2E and ToTTo) uses ROUGE-L as metrics.
    The closed-source LLMs always outperform the open-source LLMs and we report the performance difference $\Delta$ between them.
    We include the table representation (``Table Repr'') and prompting methods that yield the best performance next to the columns that report open-source and closed-source LLM scores, respectively.
    For consistency across different datasets, we \textbf{do not include the performances with highlighting cells} in this comparison.}
    \label{tab:open-vs-close}
\end{table*}

\Cref{tab:open-vs-close} provides the details of what combination of table representation and prompting method yields the best performance with respect to the Llama-70B and GPT-4 models.

\section{Prompt Examples}
\label{app-sec: prompt-examples}

\begin{figure*}[t]
    \centering
    \includegraphics[width=1\linewidth]{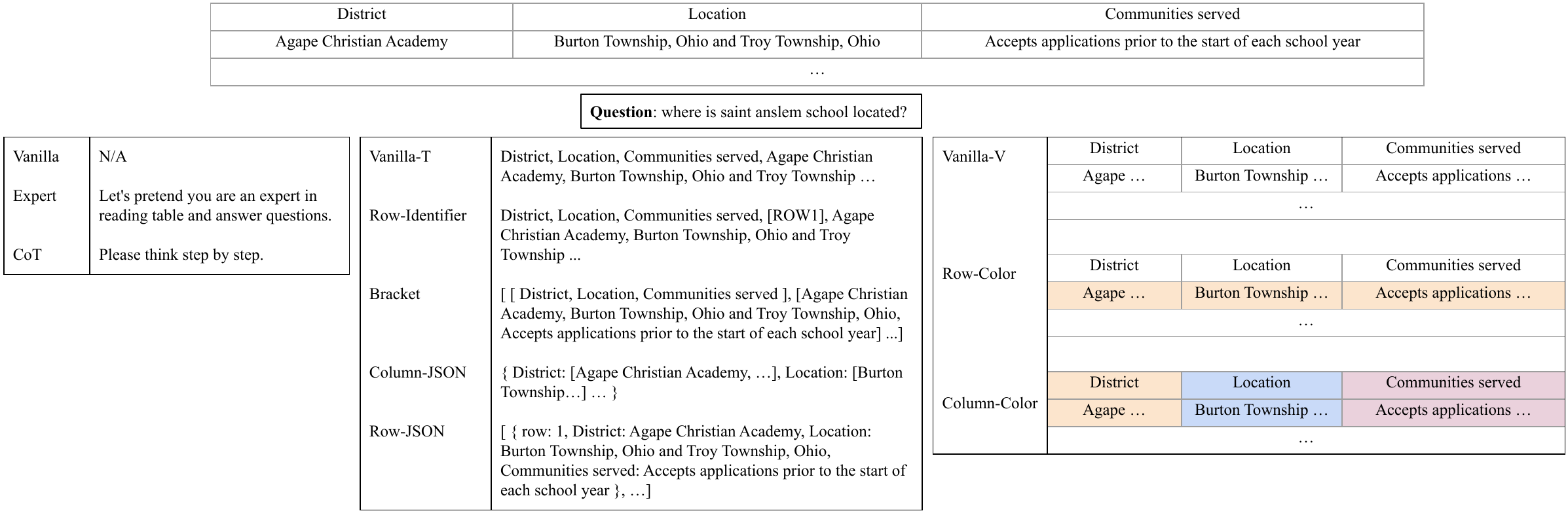}
    \caption{An example of how we construct the prompt for WikiTQ.
    Given the table and question, we choose from the three prompting methods, and combine with either the text-based or the image-based table representation.}
    \label{fig:prompt-example}
\end{figure*}

\Cref{fig:prompt-example} gives an example of how we construct our prompt for an instance in WikiTQ.

\section{LLMs' Generation Examples on Each Dataset}
\label{app-sec: dataset-examples}

\begin{figure*}[t]
    \centering
    \includegraphics[width=0.95\linewidth]{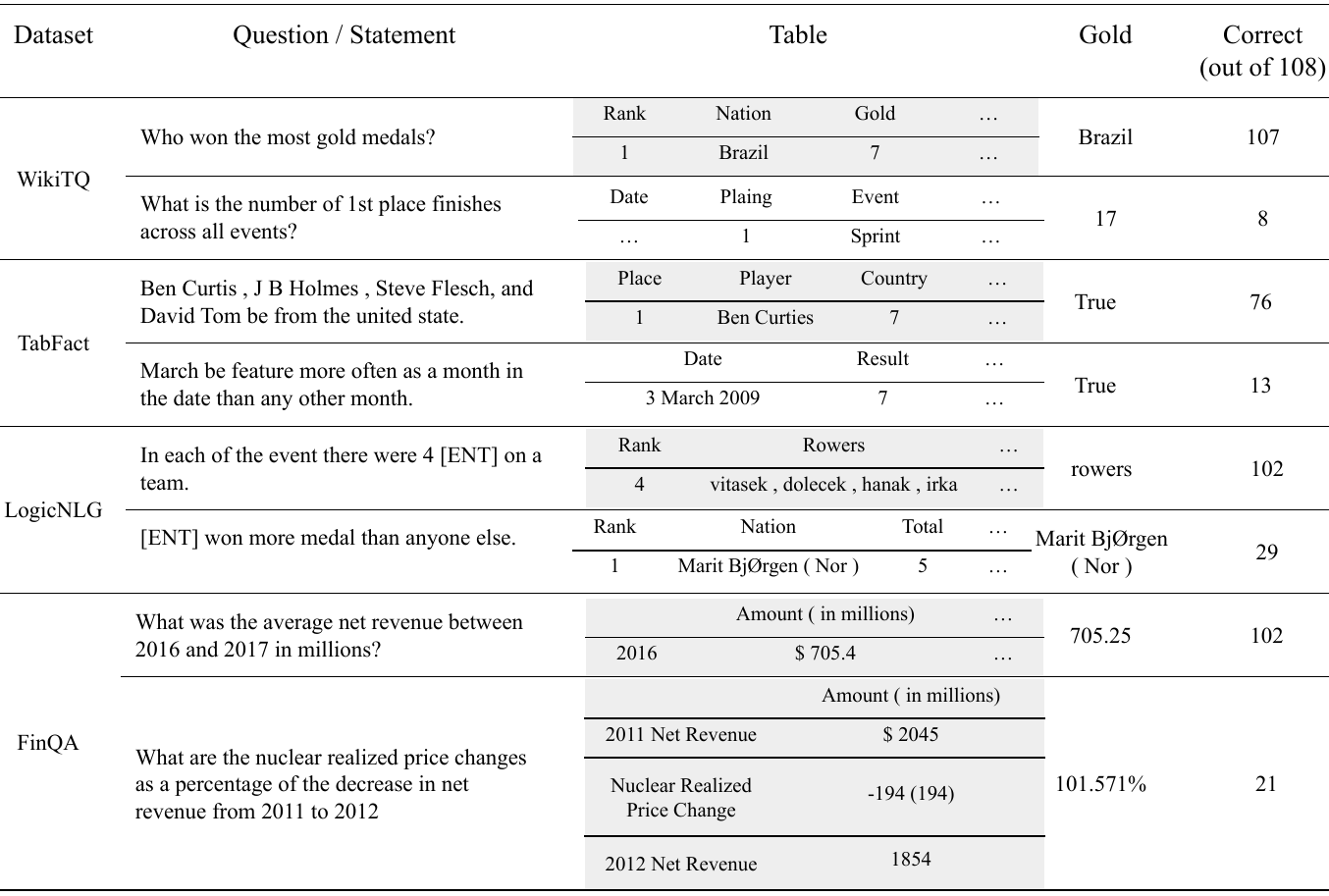}
    \caption{Examples from WikiTQ, TabFact, LogicNLG, FinQA with the number of correctly answered cases.
    For each example, we have 108 cases corresponding to the three prompting methods, five text-based table representations, and six LLMs, together with three prompting methods, three image-based table representations, and two LLMs.
    We omit some table content to assist readers.}
    \label{fig:dataset-examples}
\end{figure*}

\begin{figure*}
    \centering
    \includegraphics[width=0.95\linewidth]{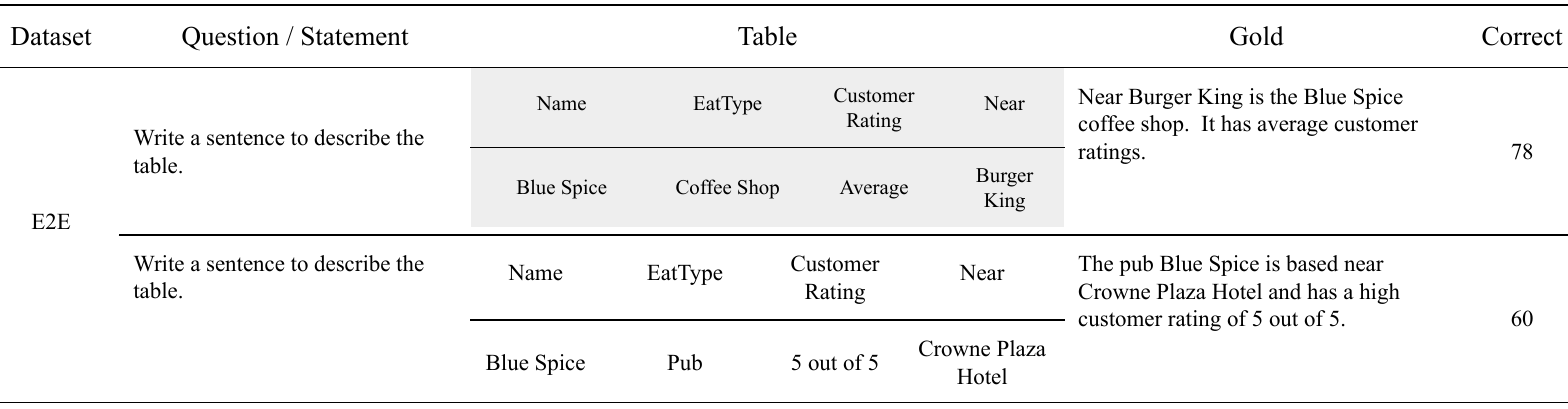}
    \caption{Examples from the E2E dataset with the number of generations that capture all the table information without any false information (manually annotated by the authors). 
    For each example, we have 102 cases as we exclude the Row-Identifier because there is one row for each table.}
    \label{fig:e2e-example}
\end{figure*}

\begin{figure*}
    \centering
    \includegraphics[width=0.95\linewidth]{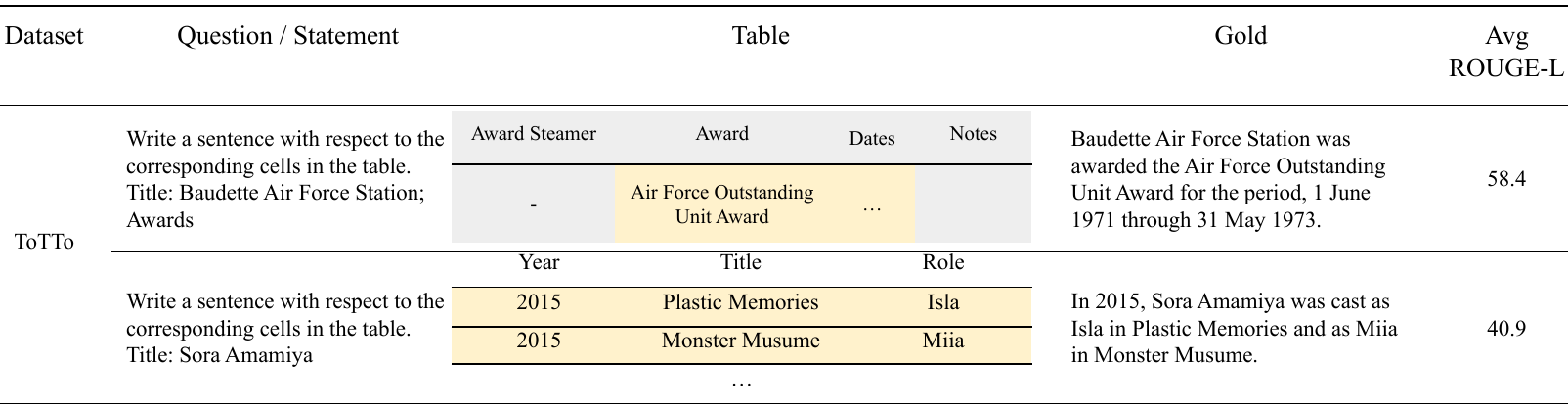}
    \caption{Examples from the ToTTo dataset with the average ROUGE L scores for the generation. 
    For each example, we have 108 cases similar to \Cref{fig:dataset-examples}.
    Since ToTTo requires to generate information about relevant cells in the table, we provide the relevant cells' information through text across all the experiments on ToTTo.}
    \label{fig:totto-example}
\end{figure*}

\Cref{fig:dataset-examples} gives examples for WikiTQA, TabFact, LogicNLG, and FinQA datasets we use, how many combinations of LLMs, table representations, and prompting techniques can answer the question correctly.
We notice that LLMs tend to answer well in general if the example focuses on extracting information from the table, but answer poorly if the question involves some arithmetic reasoning such as counting rows and comparing with others (examples from TabFact and LogicNLG), or complicated calculation that involves several steps (the example from FinQA).
\Cref{fig:e2e-example} provides examples from E2E dataset.

\Cref{fig:totto-example} provides examples from the ToTTo dataset, the models generally describe information better when there is less information.

\section{Additional Experiments on GPT-4o, Llama-3 and Gemma models}
\label{app-sec: addition-experiments}

\Cref{tab:gpt4-o-llama-3-gemma} provides additional results for vanilla prompting for GPT-4o, Llama-3 and Gemma.
We have observed that the GPT-4o performs similarly if we pass the table either through image or text.

\begin{table*}[t]
    \small
    \centering
    \begin{tabular}{lccrrccc}
    \toprule
         &  \multicolumn{2}{c}{GPT-4o}   &\multicolumn{2}{c}{Llama-3} &\multicolumn{2}{c}{Gemma} & Metric\\
         &  Vanilla-T &  Vanilla-V  & 8B & 70B  & 2B & 7B\\ 
    \midrule
    
    WikiTQ &  88.0  &  82.0 & 43.0 & 69.0 & 20.0 & 34.0& \multirow{4}{*}{Acc}\\
    TabFact &  69.7  &  70.4 & 48.5 & 71.9 & 30.2 & 41.1\\
     LogicNLG & 54.2 &  54.6 & 25.8 & 31.1 & 15.1 & 20.1 \\
     FinQA  & 71.0 &  68.0 & 44.0 & 53.0 & 2.0& 6.0 \\
     \midrule
     ToTTo & 44.0 &  44.8& 8.8 & 45.6 & 13.5 & 27.2&  
     \multirow{2}{*}{ROUGE-L}\\
       E2E & 44.2 &  43.7& 6.6 & 18.1  & 17.7 & 18.6 \\ 
    \bottomrule 
    \end{tabular}
    \caption{Results for vanilla prompting GPT4-o, Llama-3, and Gemma
   }
    \label{tab:gpt4-o-llama-3-gemma}
\end{table*}

\end{document}